\documentclass[10pt,journal,letterpaper,compsoc]{IEEEtran}
\pdfoutput=1
\usepackage[style=ieee,backend=biber]{biblatex} 
\bibliography{biblio}

\usepackage[pdftex]{graphicx}
\graphicspath{{figs/}}

\usepackage{amssymb}
\usepackage[cmex10]{amsmath}
\usepackage{algorithmic}

\usepackage{multirow}

\usepackage[tight,normalsize,sf,SF]{subfigure}

\usepackage{stfloats}
\usepackage{url}

\usepackage{hyperref}

\usepackage{todonotes}
\newtheorem{theorem}{Theorem}

\newcommand{\iid}{i.i.d.\ }
\newcommand{\eg}{e.g.\ }
\newcommand{\ie}{i.e.\ }
\newcommand{\cf}{c.f.\ }

\newcommand{\acro}[1]{\textsc{\MakeLowercase{#1}}}
\newcommand{\B}{\mathcal{B}}   
\newcommand{{\Vol}}{\mathcal{V}} 
\newcommand{\abs}[1]{\lvert#1\rvert}
\newcommand{\norm}[1]{\lVert#1\rVert}

\DeclareMathOperator{\supp}{supp}
\DeclareMathOperator{\Exp}{\mathbb{E}}  
\DeclareMathOperator*{\argmin}{arg\,min}  
\DeclareMathOperator*{\argmax}{arg\,max}  

\newcommand{\ud}{\mathrm{d}}  
\newcommand{\R}{\mathbb{R}}   
\newcommand{\M}{\mathcal{M}}  

\newcommand{\A}{\mathbf{A}}
\newcommand{\C}{\mathbf{C}}

\newcommand{\x}{\mathbf{x}}
\newcommand{\D}{\mathbf{D}}
\newcommand{\s}{\mathbf{s}}
\newcommand{\w}{\mathbf{w}}
\newcommand{\z}{\mathbf{z}}
\newcommand{\y}{\mathbf{y}}
\newcommand{\Real}{\mathbb{R}}

\newcommand{\I}{\mathbf{I}}
\newcommand{\sign}{\textrm{sign}}

\newcommand{\httpurl}[1]{\href{http://#1}{\nolinkurl{#1}}}

\begin{document}
\title{Kernels on Sample Sets \\ via Nonparametric Divergence Estimates}
%
%
%
%

\author{
    Danica~J.~Sutherland,
    Liang~Xiong, 
    Barnab\'as~P\'oczos,
    and~Jeff~Schneider%
\IEEEcompsocitemizethanks{
    \IEEEcompsocthanksitem The authors are with the School of Computer Science, Carnegie Mellon University, Pittsburgh, PA, 15213.
    }%
\thanks{}}

%
%

\markboth{}{}
%



\IEEEcompsoctitleabstractindextext{%
\begin{abstract}
Most machine learning algorithms, such as classification or regression, treat the individual data point as the object of interest.
Here we consider extending machine learning algorithms to operate on \emph{groups} of data points.
We suggest treating a group of data points as an \iid sample set from an underlying feature distribution for that group.
Our approach employs kernel machines with a kernel on \iid sample sets of vectors.
We define certain kernel functions on pairs of distributions,
and then use a nonparametric estimator to consistently estimate those functions based on sample sets. 
The projection of the estimated Gram matrix to the cone of symmetric positive semi-definite matrices enables us to use kernel machines for classification, regression, anomaly detection, and low-dimensional embedding in the space of distributions.
We present several numerical experiments both on real and simulated datasets to demonstrate the advantages of our new approach.
\end{abstract}
}


\maketitle

\IEEEdisplaynotcompsoctitleabstractindextext

%
\IEEEpeerreviewmaketitle


\section{Introduction}

Functional data analysis is a new and emerging field of statistics and machine learning.
It extends the classical multivariate methods to the case when the data points are functions \parencite{Ramsay05FDA}. In many areas, including meteorology, economics, and bioinformatics \parencite{muller08inferring}, it is more natural to assume that the data consist of functions rather than finite-dimensional vectors. This setting is especially natural when we study time series and we wish to predict the evolution of a quantity using some other quantities measured along time \parencite{kadri:nonlinear}.
Although this problem is important in many applications, the field is quite new and immature; we know very little about efficient algorithms.

Here we consider a version of the problem where the input functions are probability densities, which we cannot observe directly.
Instead, we have finite \iid samples from them.
If we can do machine learning in this scenario, then we have a way to do machine learning on groups of data points:  we treat each data point in the group as a sample point from the underlying feature distribution of the group. We can then use this approach to generalize kernel machines from finite-dimensional vector spaces to the domain of sample sets, where each set represents a distribution.

This problem of machine learning on groups of data points is important
because in many applications, the data have natural representations as sets of data, and methods for combining these sets into a single vector for use in traditional machine learning techniques can be problematic.
If we wish to find unusual galactic clusters in a large-scale sky survey
or recognize certain patterns in a turbulent vector field,
finding appropriate representations of these groups as feature vectors is a difficult problem in itself.
Similarly, images have often successfully been represented as a collection of local patches, but methods for collapsing those collections into a single feature vector present problems.

We therefore develop methods for \emph{classification and regression of sample sets}.
In the classification problem our goal is to find a map from the space of sample sets to the space of discrete objects, while in the regression problem (with scalar response) the goal is to find a map to the space of real numbers.
We choose to do so by taking advantage of the support vector machine (SVM) approach; if we can define an appropriate kernel on sample sets, algorithms for classification and scalar-response regression immediately follow.

We also show how kernel machines on the space of distributions can be used to find \emph{anomalous distributions}.
The standard anomaly/novelty detection approach only focuses on finding individual points \parencite{varun:09:ad}.
It might happen, however, that each measurement in a sample set looks normal, but the joint distribution of these values is different from those of other groups.
Our goal is to detect these anomalous sample sets/distributions.

Finally, we develop an algorithm for \emph{distribution regression}.
Here we handle a regression problem that has a distribution response: we find a linear map from the space of distributions to the space of distributions.
As an application, we show how to use this method to generalize \emph{locally linear embedding} (LLE, \parencite{roweis00LLE}) to distributions.


To generalize kernel machines to the space of distributions, we will introduce and estimate certain kernel functions of distributions.\footnote{By kernels we mean the kernel functions of a reproducing kernel Hilbert space. We do not use kernel density estimation in this paper.}
The estimators are nonparametric, consistent under certain conditions, and \emph{avoid consistent density estimation}.

Implementations of our methods and the experiments of Section~\ref{sec:Numerical} are publicly available at \httpurl{autonlab.org}.

\section{Related Work}\label{sec:related}

We will use a generalized version of the R\'enyi divergence estimator of \textcite{poczos11aistats}.
This generalization will enable us to estimate linear, polynomial, and Gaussian kernels between distributions.
There are other existing methods for divergence estimation, as well as other kinds of kernels between distributions and sets.
Below we review the most popular methods, discuss some of their drawbacks, and explain why a new approach is needed.

\paragraph*{\textbf{Parametric kernels}}
\textcite{Jebara04probabilityproduct} introduced probability product kernels on distributions.
Here a parametric family (\eg exponential family) is fitted to the densities, and these parameters are used to estimate inner products between distributions.
\textcite{Moreno04akullback-leibler} similarly use Gaussian-based models to estimate Kullback-Liebler (\acro{KL}) divergences in a kernel.
The Fisher kernel \parencite{Jaakkola98exploitinggenerative} also works on parametric families only.
In practice, however, it is rare to know that the true densities belong to these parametric families.
When this assumption does not hold, these methods introduce some unavoidable bias in estimating inner products between the densities.
In contrast, the estimator we study is completely nonparametric and provides provably consistent kernel estimations for certain kernels.
Furthermore, we avoid consistent estimation of the densities, which are nuisance parameters in our problem.

\paragraph*{\textbf{Nonparametric divergence estimation}}
\textcite{nguyen10estimating} proposed a method for $f$-divergence estimation using its ``variational characterization properties.''
This approach involves an intractable optimization over a function space of infinite dimension.
When this function space is chosen to be a reproducing kernel Hilbert space (\acro{RKHS}, \parencite{berlinet04RKHS}), this optimization problem reduces to an $m$-dimensional convex problem, where $m$ is the sample size.
This can be very demanding in practice for even just a few thousand sample points.
\textcite{Sriperumbudur10Thesis} also studies estimators that use convex or linear programming.
Our approach, which uses only $k$-\acro{NN} distances in the sample sets,
can be calculated more quickly and more simply
without needing to choose an appropriate kernel function for the \acro{RKHS}.

\textcite{2010arXiv1012.4188S} have developed $k$-nearest-neighbor based methods similar to our method for estimating general nonlinear functionals of a density. In contrast to our approach, however, their method requires $k$ to increase with the sample size $m$ and diverge to infinity. $k$th nearest-neighbor computations for large $k$ can be very computationally demanding. In our approach we fix $k$ to a small number (typically 5), and are still able to prove that the estimator is consistent.

\paragraph*{\textbf{\acro{RKHS}-based set kernels}}
There are \acro{RKHS}-based approaches for defining kernels on general unordered sets as well.
\Textcite{Kondor03akernel} introduced Bhattacharyya's measure of affinity between finite-dimensional Gaussians in a Hilbert space.
The method proposed by \textcite{Smola07ahilbert} is based on the mean evaluations of an embedding kernel between elements of the two sets, and hence its computation time is $\mathcal{O}(m^2)$.
\Textcite{muandet:measure-machines} generalize this method to operate either on distributions or on sample sets, and provide bounds on its risk deviation.
\Textcite{christmann:universal} previously showed that a similar kernel on distributions is universal and therefore SVMs based on it have good convergence properties.
A closely related method which uses pairwise Euclidean distances was proposed by \textcite{szekely04testingequal}.
Choosing an appropriate kernel function for the \acro{RKHS} can be a difficult model selection problem, however, and these methods also cannot take advantage of $k$-d trees and related methods to speed up computation as can ours.
Our empirical evaluation will also show improved results over these methods.

\paragraph*{\textbf{Quantized set kernels}}
Set kernels have been studied in computer vision domains as well.
Image classification problems often represent images as collections of local features (\cf Section~\ref{section:images}).
Perhaps the most common method for using these sets of features is to quantize the local descriptors and represent images as histograms of those features \parencite{textons}, the ``bag of words'' (\acro{BOW}) method.
\textcite{Lyu05mercerkernels} constructed composite kernels from simpler kernels defined on local features.
In the pyramid matching kernel \parencite{Grauman07pyramid}, each feature set is mapped to a multi-resolution histogram. These histogram pyramids are compared using a ``weighted histogram intersection computation.'' High-dimensional histograms can become very inefficient due to the curse of dimensionality, and selecting appropriate bin sizes is also a difficult problem. 

\paragraph*{\textbf{Nearest-neighbor set kernels}}
\textcite{nbnn} perform image classification by considering nearest-neighbor distances among local descriptors (like those used in Section~\ref{section:images}).
They argue that in the limit, their method minimizes the \acro{KL} divergence from a query image to the distribution of descriptors in the class, the optimal approach under a (dubious) Naive Bayes assumption.
\textcite{nbnn-kernel} kernelize this approach and integrate it with standard bag-of-features approaches, resulting in significant performance improvements.
Our method uses traditional object-to-object comparisons rather than object-to-class, though in principle nothing prevents us from following the approach of \cite{nbnn-kernel} in defining an object-to-class kernel, and possibly combining it with the object-to-object method described here.

\paragraph*{\textbf{Functional data analysis}}
Although the field of functional data analysis is improving quickly, there is still only very limited work available on this field.
The traditional approach in functional regression represents the functions by an expansion in some basis, e.g.\ B-spline or Fourier basis, and the main emphasis is on the inference of the coefficients \parencite{Ramsay05FDA}.
A more recent approach uses inference in an \acro{RKHS}.
For scalar responses, the first steps were made by \textcite{Cristian2007829}.
Inspired by this, \textcite{lian2007nonlinear} and \textcite{kadri:nonlinear} generalized the functional regression problem to functional responses as well.
To predict infinite-dimensional function-valued responses from functional attributes, they extended the concept of vector-valued kernels in multi-task learning \parencite{Micchelli:2005:LVF} to operator-valued kernels.
Although these methods have been developed for functional regression, they can be used for classifications of functions as well \parencite{ICML2011Kadri_509}.
We note that our studied problem is more difficult in the sense that we cannot even observe directly the inputs (densities of the distributions); only a few \iid sample points are available to us.
Luckily, in several kernel functions we do not need to know these densities; their inner product is sufficient.

\paragraph*{\textbf{Previous work}}
Our previous work \cite{poczos11uai} used a slightly less general form of the divergence estimator here for certain machine learning problems on distributions.
Our generalized method studied below can consistently estimate many divergences, including R\'enyi, \acro{KL}, $L_2$, and Hellinger distances, in a nonparametric way.
The estimator is easy to compute and does not require solving difficult kernel selection problems.

That work, however, did not investigate inner product estimation or the relation to kernel machines,
instead studying only simple $k$-NN classifiers that apply divergence estimators. 
The theoretical connection between several Hilbertian metrics and kernels has been studied by \textcite{Hein05hilbertianmetrics}, but they did not investigate the kernel estimation problem.

Instead, we will directly define inner products on distributions based on their divergences and show how to consistently estimate them.
This paper extends the results of \cite{sdm-cvpr}, which considered classification of images, and also shows how to perform anomaly detection, regression, and low-dimensional embedding based on the same techniques.


\section{Formal Problem Setting}\label{sec:setting}
Here we formally define our problems and show how kernel methods can be generalized to distributions and sample sets.
We investigate both supervised and unsupervised problems.
We assume we have $T$ inputs $X_1, \ldots, X_T$,
where the $t$th input $X_t=\{X_{t,1}, \ldots, X_{t,m_t}\}$ consists of $m_t$ \iid samples from density $p_t$.
That is, $X_t$ is a set of sample points with $X_{t,j} \sim p_t$ for $j=1,\ldots, m_t$.
Let $\mathcal{X}$ denote the set of these sample sets, so that $X_t \in \mathcal{X}$.

\subsection{Distribution classification}
In this supervised learning problem we have input-output pairs $(X_t,y_t)$
where the output domain is discrete,
\ie $Y_t \in \mathcal{Y} \doteq \{Y_1,\ldots,Y_{r}\}$.
We seek a function $f : \mathcal{X} \to \mathcal{Y}$ such that for a new input and output pair $(X,y) \in \mathcal{X} \times \mathcal{Y}$ we ideally have $f(X)=y$.

We will perform this classification with support vector machines, as reviewed here.
For simplicity, we discuss only the case when $\mathcal{Y} = \{1, -1\}$.
We can perform multiclass classification in the standard ways, \eg training either (i) $r$ one-vs-all classifiers or (ii) $r(r-1)/2$ pairwise classifiers, and then using the outputs of all classifiers to vote for the final class prediction.
In our experiments we will use the second approach.

Let $\mathcal{P}$ denote the set of density functions, $\mathcal{K}$ be a Hilbert space with inner product $\langle \cdot,\cdot\rangle_{\mathcal{K}}$, and $\phi: \mathcal{P}\to\mathcal{K}$ denote an operator that maps density functions to the feature space $\mathcal{K}$. The dual form of the ``soft margin SVM'' is \parencite{scholkopf02learning}:
\begin{align}
    \hat{\alpha} = \argmax_{\alpha \in \Real^T}
    &
    \sum_{i=1}^T \alpha_i
    - \frac{1}{2} \sum_{i,j}^T \alpha_i \alpha_j y_i y_j G_{ij}, \label{eq:SVM-dual}
    \\
    \text{s.t.} &
    \sum_{i=1}^T \alpha_i y_i = 0,
    \quad
    0 \le \alpha_i \le C
    \notag
\end{align}
where $C>0$ is a parameter
and
$G \in \Real^{T \times T}$ is the Gram matrix:
$G_{ij}\doteq
\langle \phi(p_i), \phi(p_j) \rangle_{\mathcal{K}}=K(p_i,p_j).$
The predicted class label of a test density $p$ is then
\begin{align*}
    f(p) = \sign\left( \sum_{i=1}^T \hat{\alpha}_i y_i K(p_i,p) + b \right)
\end{align*}
 where the bias term $b$ can be obtained by taking the average over all points with $\alpha_j > 0$ of $y_j-\sum_i y_i\alpha_i G_{ij}$.

There are many tools available to solve the quadratic programming task in \eqref{eq:SVM-dual},
but we still must define kernel values $\{K(p_i,p)\}_{i}$ and $\{K(p_i,p_j)\}_{i,j}$ based on the few \iid samples available to us.
We turn to that task in Section~\ref{section:methods}.

\subsection{Distribution anomaly detection}
We can use the same ideas in a one-class SVM \parencite{oneclass} to find anomalous distributions in an unsupervised manner.
To do so, we train on a set of samples considered ``normal'' (usually sampled randomly from a larger dataset), and estimate a function $f$ which is positive on a region around the training set and negative elsewhere
in a problem similar to \eqref{eq:SVM-dual}.

\subsection{Distribution regression with scalar response (DRSR)}
\label{sec:drsr}
We also sometimes wish to solve a supervised task where the output domain is continuous: a regression task on sample sets with scalar response.
For example, we may wish to approximate real-valued functionals of distributions such as entropy or mutual information, based on training points obtained by a reliable but computationally intensive Monte Carlo method.
Alternatively, we may wish to consider images as distributions and obtain the number of pedestrians crossing a street or the size of a tumor.
We can then use the same Gram matrix defined above in the support vector regression equations \parencite{scholkopf02learning} to estimate an unknown function $f : \mathcal{X} \to \R$.

\subsection{Distribution regression with distribution response (DRDR) and locally linear embedding (LLE)}
\label{sec:drdr-lle}
We can also consider the regression problem where $\mathcal{Y} = \mathcal{X}$, that is, the outputs are also \iid samples from distributions, and we are looking for a function $f: \mathcal{X} \to \mathcal{X}$.
Below we show how the coefficients of the linear regression can be calculated after transforming the distributions to the Hilbert space $\mathcal{K}$.
The regression problem is given by the following quadratic program:
\begin{align*}
    \hat{\alpha}
    &= \argmin_{\alpha \in \R^T} \, \norm{ \phi(p)-\sum_{i=1}^T \alpha_i \phi(p_i) } ^2_{\mathcal{K}} \\
    &= \argmin_{\alpha \in \R^T}
        \left[
            K(p,p) - 2\sum_{i=1}^T \alpha_i K(p_i,p)
            + \sum_{i, j}^T \alpha_i \alpha_j G_{ij}
        \right].
\end{align*}

LLE \parencite{roweis00LLE} performs nonlinear dimensionality reduction by computing a low-dimensional, neighborhood-preserving embedding of high- (but finite-) dimensional data. As an application of DRDR, we show how to generalize LLE to the space of distributions and to \iid sample sets.
Our goal is to find a map $f: \mathcal{X}\to \R^d$ that preserves the local geometry of the distributions.
To characterize this local geometry, we reconstruct each distribution from its $\kappa$ neighbor distributions by DRDR.



As above, let $X_1,\ldots,X_T$ be our training set. The squared Euclidean distance between $\phi(p_i)$ and $\phi(p_j)$ is given by
$\langle \phi(p_i)-\phi(p_j), \phi(p_i)-\phi(p_j)\rangle_{\mathcal{K}}=K(p_i,p_i)+K(p_j,p_j)-2K(p_i,p_j)$.
Let $\mathcal{N}_i$ denote the set of the $\kappa$ nearest neighbors of distribution $p_i$ among $\{p_j\}_{j\neq i}$.
The intrinsic local geometric properties of distributions $\{p_i\}_{i=1}^T$ are characterized by the reconstruction weights $\{W_{i,j}\}_{i,j=1}^T$ in the equation below:
\begin{align}
    \widehat{W} = \argmin_{W \in \R^{T\times T}}
    & \sum_{i=1}^T \norm{
        \phi(p_i) - \sum_{j \in \mathcal{N}_i} W_{i,j} \phi(p_j)
      }^2_{\mathcal{K}}
    \label{eq:LLE-weights1} \\
    \text{s.t.} &
      \sum_{j \in \mathcal{N}_i} W_{i,j} = 1,
      \; \textrm{and} \;
      W_{i,j}=0 \textrm{ if } j \notin \mathcal{N}_i. \nonumber
\end{align}
Note that the cost function in~\eqref{eq:LLE-weights1} can be rewritten as
\begin{align*}
\sum_{i=1}^T \left(G_{ii} -2\sum_{j\in\mathcal{N}_i}W_{i,j}G_{ij}
+\sum_{j\in\mathcal{N}_i}\sum_{k\in\mathcal{N}_i} W_{i,j} W_{k,j}G_{jk}\right).
\end{align*}
Having calculated the weights $\{\widehat{W}_{i,j}\}$, we compute $Y_i=f(p_i) \in \R^d$ (the embedded distributions) as the vectors best reconstructed locally by these weights:
 \begin{align*}
    \widehat{Y} =
    \argmin_{\{Y_i \in \R^d\}_{i=1}^T} \,
        \sum_{i=1}^T \, \norm{Y_i-\sum_{j\in\mathcal{N}_i} \widehat{W}_{i,j}Y_j}^2
\end{align*}
Finally, $\widehat{Y}=\{\widehat{Y}_1,\ldots,\widehat{Y}_T\}$, and $\widehat{Y}_i$ corresponds to the $d$-dimensional image of distribution $p_i$.

Note that many other nonlinear dimensionality reduction algorithms --- including stochastic neighbor embedding \parencites{Hinton02stochasticneighbor,t-sne}, multidimensional scaling \parencite{borg05MDS}, and isomap \parencite{tenenbaum00isomap} --- use only Euclidean distances between input points, and so can be generalized to distributions by simply replacing the finite-dimensional Euclidean metric with $\langle p-q, p-q\rangle^{1/2}_{\mathcal{K}} $. 

\section{Methods} \label{section:methods}

All of these methods require a kernel defined on sample sets.
We choose to define a kernel on distributions and then estimate its value using our samples.

\subsection{Distribution Kernels} \label{section:Constructing-Kernels}
Many kernels, \ie positive semi-definite functionals of $p$ and $q$,
can be constructed from the following form:
\begin{equation}
    D_{\alpha,\beta}(p\|q) =
    \int p^\alpha(x)\, q^{\beta}(x)\, p(x)\,\ud x,  \label{eq:Dalpha-beta}
\end{equation}
where $\alpha, \beta \in \R$.
Some examples are linear, polynomial,
and Gaussian kernels
$\exp\left( - \frac{1}{2} \mu^2(p, q) / \sigma^2 \right)$.
Normally, $\mu^2(p, q)$ is the $L_2$ distance between $p$ and $q$.
We can also try to use other ``distances'' here,
such as the Hellinger distance, the R\'enyi-$\alpha$ divergence,
or the \acro{KL} divergence (the $\alpha \to 1$ limit of the R\'enyi-$\alpha$ divergence).
These latter divergences are not symmetric, do not satisfy the triangle inequality, and do not lead to positive semi-definite Gram matrices.
We will address this problem in Section~\ref{sec:psd-proj}.
Table~\ref{tab:kernel-forms} shows the forms of these kernels and distances.

\begin{table}[h]
    \centering
    \caption{Kernels and squared distances we can construct with \eqref{eq:Dalpha-beta}.}
    \label{tab:kernel-forms}
    \renewcommand{\arraystretch}{1.3}
    \begin{tabular}{ccc}
        Linear kernel & $\int p q$ & $D_{0,1}$ \\
        Polynomial kernel & $\left(c + \int p q \right)^s$ & $\left(c + D_{0,1}\right)^s$ \\
        $L_2$ distance & $\int (p - q)^2$ & $D_{1,0} - 2 D_{0,1} + D_{-1,2}$ \\
        Hellinger distance & $1 - \int \sqrt{p q}$ & $1 - D_{-1/2,1/2}$ \\
        R\'enyi-$\alpha$ divergence
        & $\left(\frac{\log \int p^\alpha q^{1-\alpha}}{\alpha - 1} \right)^2$
        & $\left(\frac{\log D_{\alpha-1,1-\alpha}}{\alpha - 1} \right)^2$
    \end{tabular}
\end{table}

\subsection{Kernel Estimation} \label{section:Kernel-estimation}
We will estimate these kernel values on our sample sets
by estimating $D_{\alpha,\beta}(p\|q)$ terms for some $\alpha, \beta$.
Using the tools that have been applied for the estimation of R\'enyi entropy \parencite{Leonenko-Pronzato-Savani2008},
\acro{KL} divergence \parencite{Wang-Kulkarni-Verdu2009},
and R\'enyi divergence \parencite{poczos11aistats},
we show how to estimate $D_{\alpha,\beta}(p\|q)$ in an efficient, nonparametric, and consistent way.

Let $X_{1:n} \doteq (X_1,\ldots,X_n)$ be an \iid sample from a distribution with density $p$,
and similarly let $Y_{1:m}\doteq(Y_1,\ldots,Y_m)$ be an \iid sample from a
distribution having density $q$.
Let $\rho_k(i)$ denote the Euclidean distance of the $k$th nearest
neighbor of $X_i$ in the sample $X_{1:n} \setminus \{ X_i \}$,
and similarly let $\nu_k(i)$ denote the distance of the $k$th nearest neighbor of $X_i$ in the sample $Y_{1:m}$.

The following estimator is provably $L_2$ consistent
if $k > 2 \max(\abs{\alpha}, \abs{\beta}) + 1$ and
under certain conditions on the density
(see the Appendix for details):
 \begin{equation}
    \widehat{D}_{\alpha,\beta}
      = \frac{B_{k,d,\alpha,\beta}}{n (n-1)^\alpha m^\beta}
        \sum_{i=1}^n \rho_k^{-d\alpha}(i) \; \nu_k^{-d\beta}(i), \label{eq:D_hat}
\end{equation}
where $B_{k,d,\alpha,\beta} \doteq \bar{c}_d^{-\alpha-\beta} \, \frac{\Gamma(k)^2}{\Gamma(k-\alpha) \, \Gamma(k-\beta)}$
and
$\bar{c}_d$ denotes the volume of a $d$-dimensional unit ball.

\subsection{Positive Semi-definite Projection} \label{sec:psd-proj}

$\widehat D_{\alpha,\beta}$ is a consistent estimator of $D_{\alpha,\beta}$ under appropriate conditions, and thus plugging these estimators into the formulae in Section~\ref{section:Constructing-Kernels} gives consistent estimators for those kernels. The quality of kernel estimation therefore improves as the number of sample points increases. This does not, however, guarantee that the estimated Gram matrix is positive semi-definite (\acro{PSD}).
When we use R\'enyi or other divergences instead of the Euclidean metric in the Gaussian kernel, the resulting functional itself may be asymmetric or indefinite.
Additionally, estimation error may lead to a non-\acro{PSD} result for any kernel.
Therefore, we symmetrize the estimated Gram matrix with $(G + G^T)/2$ and then project it to the nearest symmetric \acro{PSD} matrix (in Frobenius norm) by discarding negative eigenvalues from its spectrum \cite{Higham2002}.

There are various other approaches in the literature to solve this problem.
Many use an approach similar to ours, or shift all the eigenvalues by an amount large enough to make them all positive, or flip negative eigenvalues to be positive \parencite{chen:indefinite-training}.
\Textcite{Moreno04akullback-leibler} shift and scale the divergences with $e^{-A D + B}$, but do not give details on how they choose $A$ or $B$.

\Textcite{luss:indefinite-svm} instead view the indefinite kernel matrix as a noisy approximation of a \acro{PSD} matrix, and define a single optimization problem to jointly find the ``true'' kernel and the support vectors.
\Textcite{ying:indefinite} and \textcite{chen:indefinite-training} improve the training algorithm in that setting, but the latter's empirical work found only small improvements over our ``denoise'' approach of discarding negative eigenvalues.

It is also possible to directly use an indefinite kernel in an SVM-like learning problem.
\Textcite{haasdonk:indefinite} provides a geometric interpretation as separating convex hulls in a pseudo-Euclidean space isometric to the kernel and gives some learning approaches.
\Textcite{ong:nonpositive} associate such indefinite Gram matrices with a Reproducing Kernel Kre\u{\i}n space, where an analog of the representer theorem holds;
\textcite{loosli:nonpositive} give a reformulation of the Karush-Kuhn-Tucker conditions to provide more efficient learning for this problem.
It is possible that this approach would yield better results, but these techniques remain not very well understood.

The projection approach yields a transductive learning algorithm.
If we have $T$ training and $N$ test examples and project the estimated Gram matrix $\widehat{G} \in \Real^{(T+N) \times (T+N)}$ onto the cone of \acro{PSD} matrices, this symmetric \acro{PSD} matrix defines a valid kernel on the training and test points, and thus the representer theorem holds on these points.
It might not, however, be a valid kernel on other points not in the training and test sets.
Note also that in classification and regression we need to solve the quadratic problem \eqref{eq:SVM-dual} in the training phase, using the Gram matrix of the training points only. It can be crucial for this training Gram matrix to be \acro{PSD} for \acro{QP} solvers to find a solution of \eqref{eq:SVM-dual}.

In practice, however, we can also use this approach inductively by predicting based on unprojected kernel estimates. Although this approach is not supported by the representer theorem, it is computationally well-defined. The two approaches perform identically in the experiment of Section~\ref{section:USPS}, and we also see good results with the inductive approach in Section~\ref{section:turbulence}.

\subsection{Algorithm Summary}
Fig.~\ref{alg:classification} gives pseudocode for classification.
\begin{figure}[h]
\begin{algorithmic}
    \STATE {\bfseries Input:}
        $T$ training and $N$ testing feature sets \\
        $\qquad X_t = \{X_{t,1}, \ldots, X_{t,m_t} \}$ for $t = 1, \dots, T+N$; \\
        $\quad T$ training class labels $Y_{t} \in \{0,1\}$ for $t = 1, \dots, T$.
    \STATE
    \FOR{$i,j \in \{ 1, \dots, N+T \}^2$}
    \STATE Estimate $D_{\alpha,\beta}(p_i\|p_j)$ by $\widehat{D}_{\alpha,\beta}(X_{i}\|X_{j})$ \eqref{eq:D_hat}.
        \STATE Calculate the kernel estimate $\widehat{G}_{i,j}$ from this estimate.
    \ENDFOR
    \STATE Symmetrize the Gram matrix: $\widehat{G}:=(\widehat{G}+\widehat{G}^T)/2$.
    \STATE Discard negative eigenvalues from the spectrum of $\widehat{G}$.
    \STATE Solve \eqref{eq:SVM-dual} and calculate $\hat{\alpha}, b$ in the SVM.
    \STATE
    \FOR{$j = T + 1, \dots, T + N$}
        \STATE Predict $\hat{y}_j = \sign(\sum_{i=1}^T \hat\alpha_i y_i \widehat{G}_{i,j} + b)$.
    \ENDFOR
\end{algorithmic}
   \caption{Algorithm summary for distribution classification.}
   \label{alg:classification}
\end{figure}

\section{Numerical Experiments} \label{sec:Numerical}

We now demonstrate the applicability of our method in several numerical experiments. As mentioned previously, the code and datasets used in this section are publicly available at \httpurl{autonlab.org}.

%

\subsection{USPS Classification}\label{section:USPS}
The USPS dataset%
\footnote{\httpurl{www-stat-class.stanford.edu/~tibs/ElemStatLearn/data.html}}
\parencite{hull94usps} consists of 10 classes; each
data point is a $16 \times 16$ grayscale image of a handwritten digit. The human
classification error rate is 2.5\%. The best algorithms, \eg the Tangent Distance algorithm \parencite{simard98tangentdistance}, achieve this error rate.



This dataset is simple enough that raw images can be used as features for classifications, and even the Euclidean distances between images have high discriminative values. We can easily construct a considerably more difficult noisy version of this dataset, however, in which inner products and Euclidean distances become much less useful. Those measures disregard the proximity of pixels to one another. As resolutions increase, that information becomes more and more necessary.

\begin{figure}[th] 
    \centering
    \includegraphics[width=.49\textwidth]{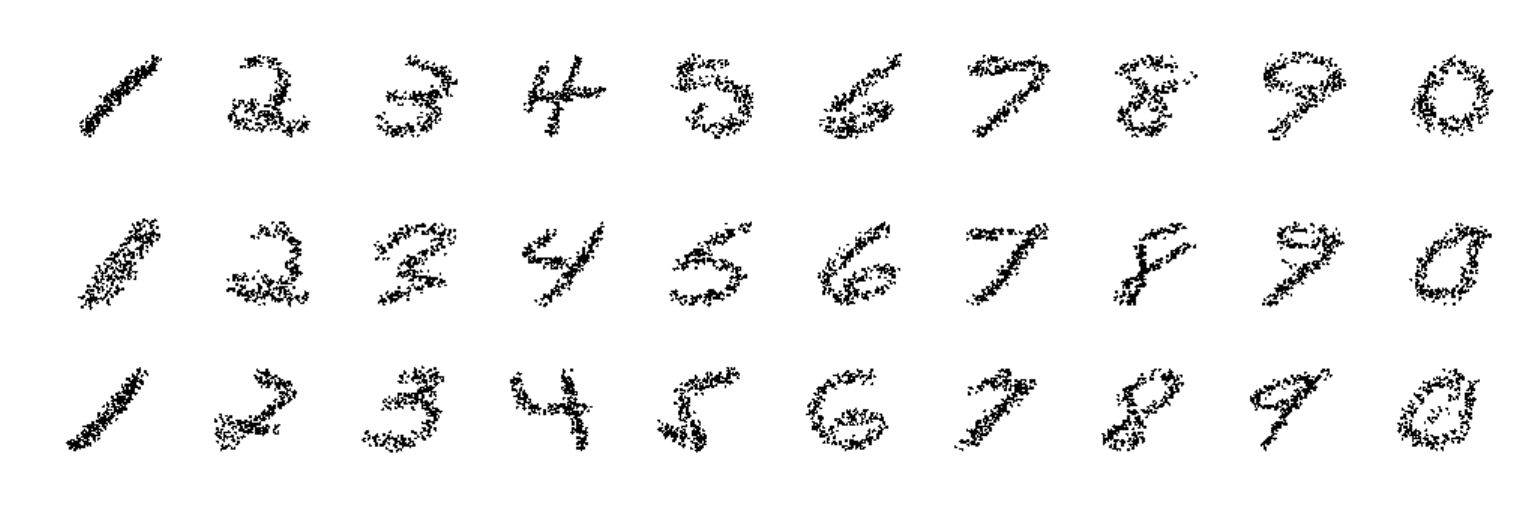}
    \caption{Noisy USPS dataset.} \label{fig:NoisyUSPSdataset}
\end{figure}

We resized the images to $160 \times 160$ and normalized pixel intensities to sum to 1 for each image. We then treat each image as a probability distribution, where a 2d coordinate's sampling probability is proportional to its (negated) grayscale value. We drew an \iid sample of size 500 from each distribution and added Gaussian noise (zero mean, 0.1 variance) to each. Example noisy images are shown in Fig.~\ref{fig:NoisyUSPSdataset}.

We took 200 such samples from each class of the noisy dataset and performed 16 runs of two-fold cross-validation, so that each 10-class SVM (\ie 45 pairwise SVMs) had 1000 testing and 1000 training points.

We tuned parameters in a grid search using 3-fold cross-validation on the training set, randomly picking among the configurations with the best performance.
The margin penalty $C$ was selected from the set $\left\{ 2^{-9}, 2^{-6}, \dots, 2^{21} \right\}$;
the degree of polynomial kernels (both homogeneous and inhomogeneous) was selected from $\left\{ 1, 2, 3, 5, 7, 9 \right\}$,
while Gaussian kernels were tried with kernel widths $\sigma$ from $\left\{ 2^{-4}, 2^{-2}, \dots, 2^{10} \right\}$ times the median of the nonzero elements of the Gram matrix.

The mean accuracy for this 10-class problem was $93\%$ for Gaussian kernels on the original images using pixel values as features, but only $83\%$ on the noisy data.
Polynomial kernels performed similarly.
Using our estimates of the R\'enyi-.9 divergence with $k = 5$ nearest neighbors, however, we obtained $96\%$ accuracy on the noisy dataset with both transductive and inductive evaluation, approaching the human error rate.
Results for other choices of $\alpha$s in R\'enyi divergence were indistinguishable, while Hellinger, $L_2$, and polynomial kernels performed slightly worse.
Table~\ref{table:usps-full-results} presents means and standard deviations of all the accuracies.
``Raw'' means that image pixels were used directly as features in computing inner products; ``NP'', standing for ``nonparametric,'' means kernel values were estimated using \eqref{eq:D_hat}.

\begin{table}[ht]
    \centering
    \caption{%
        Mean and std classification accuracies for various methods.
        See the text for descriptions.
    }
    \label{table:usps-full-results}
    \renewcommand{\arraystretch}{1.3}
    \begin{tabular}{c|cc|cc}
        & & & Transductive & Inductive \\
        \hline
        \multirow{2}{*}{orig} & \multirow{2}{*}{raw} & Polynomial & \multicolumn{2}{c}{$93.5 \pm .5$} \\
        && $L_2$ & \multicolumn{2}{c}{$93.3 \pm .6$} \\
        \hline
        \multirow{10}{*}{noisy} & \multirow{2}{*}{raw} & Polynomial & \multicolumn{2}{c}{$82.1 \pm .5$} \\
        & & $L_2$ & \multicolumn{2}{c}{$83.4 \pm .4$} \\
        \cline{2-5}
        & \multirow{8}{*}{NP} & Polynomial & $92.7 \pm .5$ & $92.9 \pm .5$ \\
        & & $L_2$  & $93.7 \pm .3$ & $93.7 \pm .3$ \\
        & & R\'enyi-.2 & $95.8 \pm .3$ & $95.8 \pm .3$ \\
        & & R\'enyi-.5 & $95.9 \pm .3$ & $95.8 \pm .3$ \\
        & & R\'enyi-.8 & $96.1 \pm .3$ & $96.0 \pm .2$ \\
        & & R\'enyi-.9 & $96.0 \pm .3$ & $96.0 \pm .3$ \\
        & & R\'enyi-.99 & $96.0 \pm .4$ & $96.1 \pm .4$ \\
        & & Hellinger & $94.9 \pm .4$ & $94.9 \pm .3$ \\
    \end{tabular}
\end{table}

\begin{figure}  [h] 
    \centering
    \subfigure[using raw images]{
        \includegraphics[width = 1.5in]{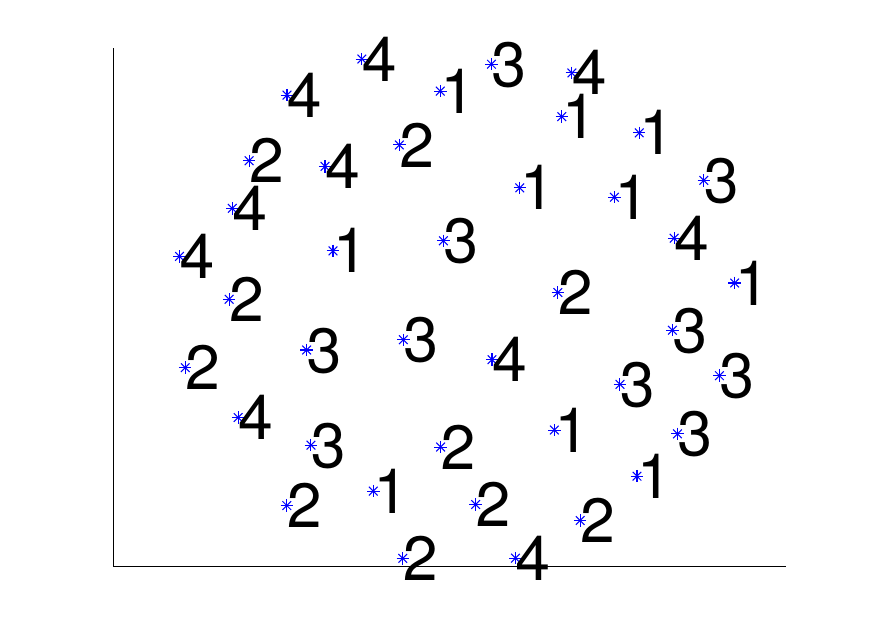} \label{fig:MDS-imdifference}}
    \subfigure[using distributions]{
        \includegraphics[width = 1.5in]{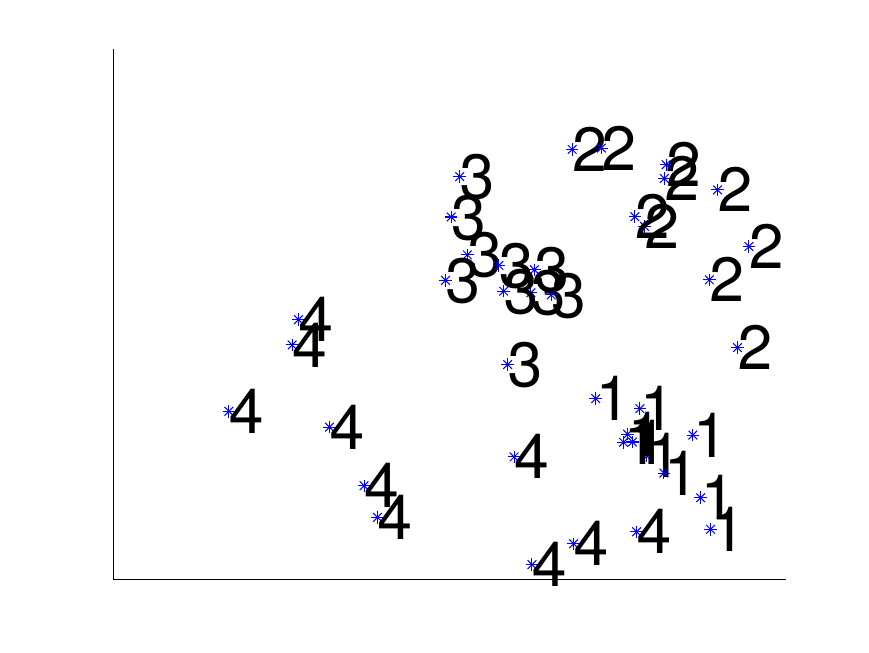} \label{fig:MDS-distribution}}
    \caption{Multidimensional scaling using (a) Euclidean metric between the images, and (b) the estimated Euclidean distance between the distributions of the coordinates of the image pixels.}
\end{figure}

To demonstrate visually that our estimated kernel evaluations give more information about the structure of the noisy dataset than do Euclidean distances between raw images, we performed multidimensional scaling to 2d using ten instances of the digits $\{1,2,3,4\}$. Fig.~\ref{fig:MDS-imdifference} shows the embedding using Euclidean distances between images and Fig.~\ref{fig:MDS-distribution} the embedding using our estimate of the Euclidean distance between samples. We see that our method was able to preserve the class structure of the dataset. The letters form natural clusters, as opposed to the raw Euclidean distance, where the scaling is uninformative. This helps explain why performance was so much better with the distribution-based kernels.

\subsection{Natural Image Classification} \label{section:images}

We now turn to the use of our kernel for classification tasks in computer vision, in particular whole-image classification of objects and scenes.
This section extends the results of \textcite{sdm-cvpr}.

One popular set of algorithms for image classification is based on the
``bag of words'' (\acro{BoW}) representation.
With \acro{BoW}, each image is considered as a collection of \emph{visual words}, so that an image of the coast might contain visual words describing patches of sky, sea and so on --- though the words are not labeled and may not correspond to recognizable concepts.
The locations of and dependencies among those visual words are ignored.
The collection of unique visual words is called the \emph{visual vocabulary}.

Given a set of images, the visual words and vocabulary are often constructed as follows:
\begin{enumerate}
    \item Select local patches from each image.
    \item Extract a feature vector for each patch. \label{item:patch}
    \item Cluster/quantize all the patch features into $V$ categories.
By doing this, each category of patches would represent an area with similar extracted features, hopefully a visual concept such as ``sky'' or ``window.''
These $V$ categories form the visual vocabulary, and each patch, represented by its category number, becomes a visual word.
    \item Within each image, count the number of different visual words and construct a histogram of size $V$. This is the \acro{BoW} representation of the image.
\end{enumerate}

The \acro{BoW} representation is then typically used in kernels based on the chi-squared distance between histograms \parencite{zhang:local-features}, or on Euclidean distance between histograms discounted by a square-root transformation \parencite{perronnin:large-img-clf}, which are similar.
Those transformations are probably successful in part because they approximate a model that does not make the unrealistic \iid assumption for features \parencite{cinbis:non-iid}.

Rather than performing the quantization steps, however, we represent each image directly as the set of extracted features from each patch (\ie stopping at step (\ref{item:patch}) above; we refer to this as the ``bag of features'' representation, \acro{BoF}).
We can then apply our divergence-based kernels, or any other group kernel, to the images.


\subsubsection*{Feature Extraction} \label{sec:exp_img_data}
We extract dense \acro{SIFT} features \cite{sift} using the techniques of \textcite{bosch:08:bow}.
For each image, we compute \acro{SIFT} descriptors at points on a regular grid with a step size of $20$ pixels.
In an attempt toward scale invariance, at each point we compute several \acro{SIFT} descriptors, each of which is $128$-dimensional, with a range of radii.
After the feature extraction, each image is represented by a variable number of 128-dimensional feature vectors.
We can also include color information in the \acro{SIFT} features by converting the images to \acro{HSV} color space and computing \acro{SIFT} features independently in each color channel \parencite{bosch:08:bow}.
\acro{SIFT} features at the same location with the same bin size are then concatenated together to produce a ``color \acro{SIFT}'' feature vector, with dimensionality 384.
We use \acro{PCA} to reduce the feature vectors' dimensionality for computational expedience, and finally normalize each dimension to have zero mean and unit variance.
We used the \textsc{VLFeat} package \parencite{vlfeat} and its
\acro{PHOW} functionality to extract the SIFT features.

\subsubsection*{Performance Evaluation}

We evaluated several kernels for image classification.
Parameter tuning for the SVM's margin penalty $C$ and, when applicable, the Gaussian kernel width $\sigma$ is performed as for the USPS dataset (Section~\ref{section:USPS}).
All kernel matrices are projected to be symmetric \acro{PSD} before use.

\paragraph*{{\bf Nonparametric divergence kernels}}
These kernels are based on the proposed nonparametric R\'enyi-$\alpha$ divergence estimators (\textbf{\acro{NPR}-$\alpha$})
and Hellinger distance estimators (\textbf{\acro{NPH}}).
In this high-dimensional setting, estimation of polynomial kernels and $L_2$ distance is more difficult and did not produce reliable results; the estimate of $\int \! p q$ in particular was extremely poor.
We use the $k = 5$th nearest neighbors here.
For \acro{NPR}, we generally test the performance with $\alpha \in \left\{ .5, .8, .9, .99 \right\}$.
Note that as $\alpha \to 1$ the R\'enyi-$\alpha$ divergence approximates the \acro{KL} divergence,
and when $\alpha = .5$ it is twice the Bhattacharyya distance.

\paragraph*{{\bf Parametric kernels}}
These kernels assume the data follows a Gaussian distribution or a mixture of Gaussian distributions.
We first fit the densities and then compute either the $\acro{KL}$ divergence between groups (\textbf{\acro{G-KL}, \acro{GMM-KL}}) \parencite{Moreno04akullback-leibler}
or \emph{probability product kernels} with $\alpha = .5$ (\textbf{\acro{G-PPK}, \acro{GMM-PPK}}), which estimate the Bhattacharyya coefficients between the Gaussians \parencite{Jebara04probabilityproduct}.
We use three components in our \acro{GMM}s.
\acro{GMM-KL} has no analytic form, so we use a Monte Carlo approximation with 500 samples.

\paragraph*{{\bf \acro{BoW} kernels}}
We used \textbf{\acro{BoW}} kernels as described previously, using the chi-square distance between histograms.
We used \emph{k-means} for quantization with a vocabulary size (number of clusters) of 1000 for color images and 500 for grayscale images.
We also considered a kernel based on Euclidean distances between histograms processed by \textbf{\acro{PLSA}} using 25 topics \parencite{plsa}.
Note that in both cases, the features are quantized based on the original feature vectors, \emph{not} the features after \acro{PCA}.

\paragraph*{{\bf Pyramid matching kernel}}
We also test the vocabulary-guided pyramid matching kernel \textbf{\acro{PMK}} \parencite{grauman:06:vgpmk},
the preferred \acro{PMK} variant for high-dimensional data.
We used the authors' implementation \emph{libpmk}\footnote{\httpurl{people.csail.mit.edu/jjl/libpmk}} with their recommended parameters.

\paragraph*{{\bf Mean map kernel}}
We also consider the \emph{mean map kernel} \textbf{\acro{MMK}} \parencite{Smola07ahilbert}, also known as the \emph{mean match kernel} \parencite{Lyu05mercerkernels} to the computer vision community. The MMK between two groups of vectors ${\bf X}=\{x_1,\ldots,x_m\}$ and ${\bf Y}=\{y_1,\ldots,y_n\}$ is defined as $k_{MM}({\bf X},{\bf Y})=\frac{1}{mn}\sum_{i=1,j=1}^{m,n} k(x_i,y_j)$. In other words, \acro{MMK} is the average kernel matching score between every pair of points between the two groups. We let the point-wise matching kernel be the Gaussian kernel $k(x,y)=\exp{\left(-\left\| x-y \right\|_2^2/\sigma^2\right)}$, where the kernel width $\sigma$ is tuned in the same way as other parameters. To avoid the high computational cost of \acro{MMK} ($O(mn)$ for each pair of groups), we randomly choose at most $500$ points from each group to get \acro{MMK}, so that the computation is affordable while the approximation error is small.

In \parencite{Lyu05mercerkernels}, it was argued that in \acro{MMK}, good point-wise matches will be ``swamped'' when averaged with a larger number of bad matches. The author then proposed to exponentiate the point-wise kernels so that good matches (larger kernel values) will dominate the average. In our case of Gaussian kernels, the exponentiation is subsumed into the kernel width and will be selected by cross-validation.

\subsubsection{Object Recognition}
\begin{figure}[htb]
  \centering
  \includegraphics[width=1.3cm]{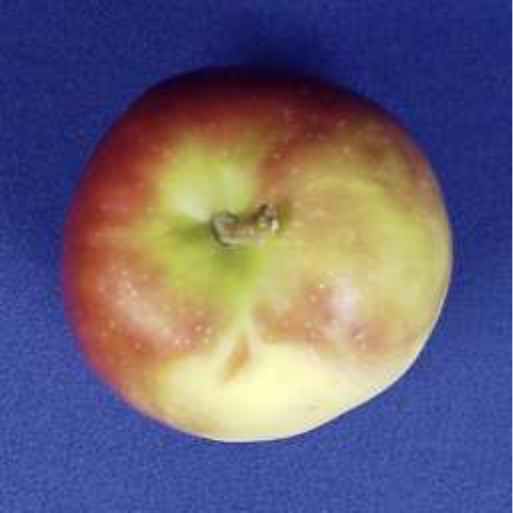}
  \includegraphics[width=1.3cm]{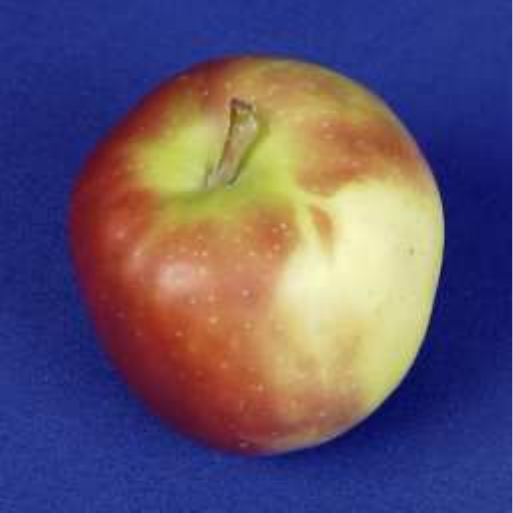}
  \includegraphics[width=1.3cm]{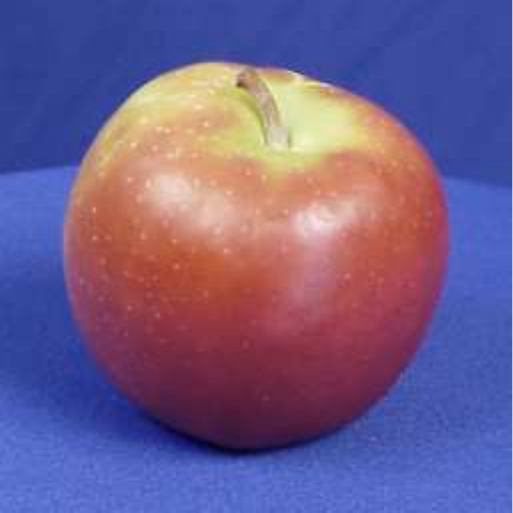}
  \includegraphics[width=1.3cm]{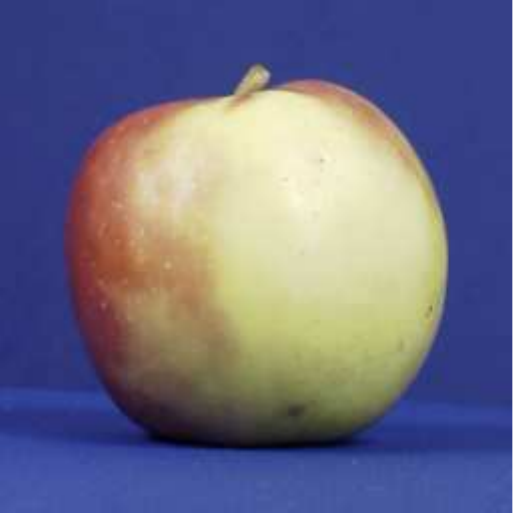}
  \includegraphics[width=1.3cm]{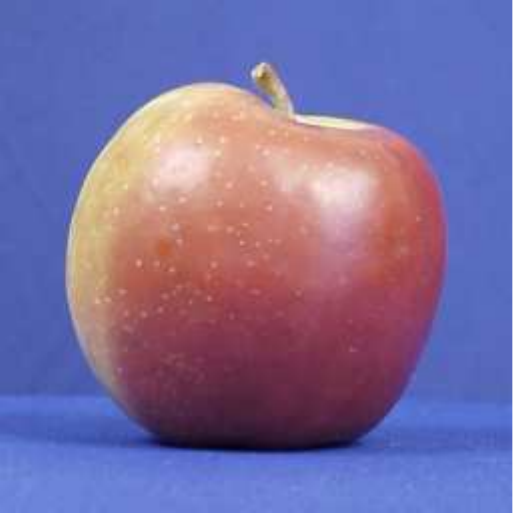} \\
  \includegraphics[width=1.3cm]{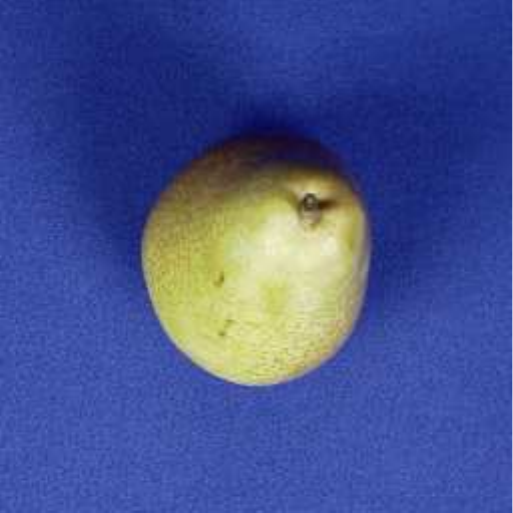}
  \includegraphics[width=1.3cm]{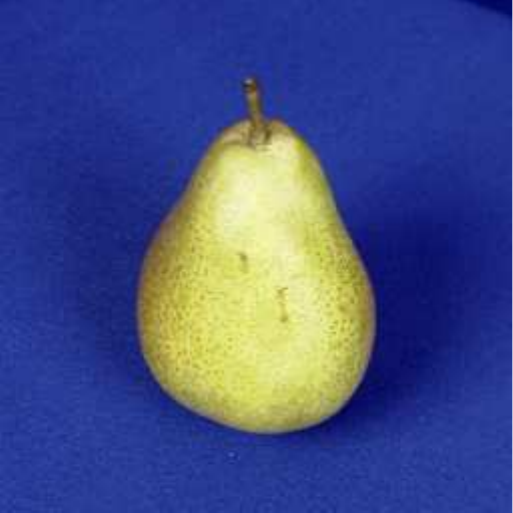}
  \includegraphics[width=1.3cm]{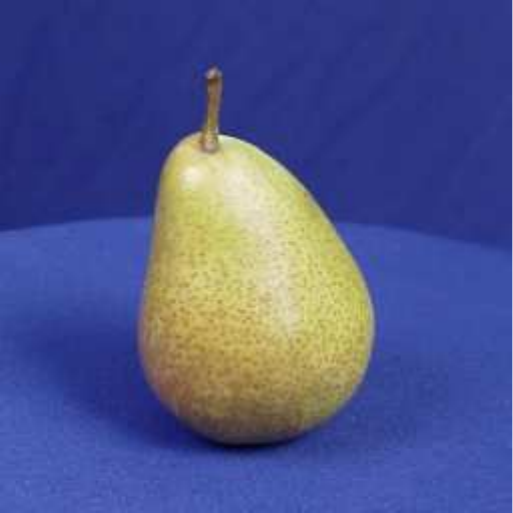}
  \includegraphics[width=1.3cm]{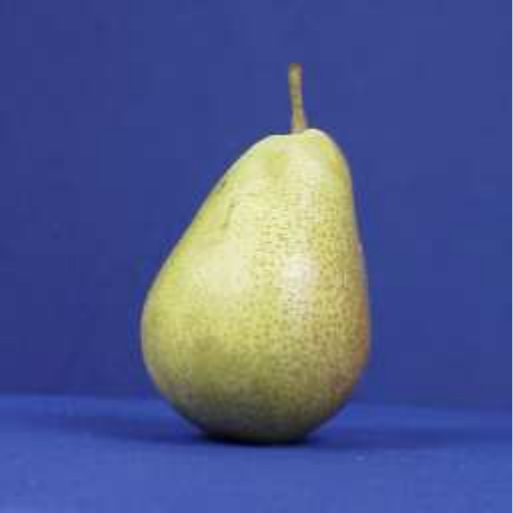}
  \includegraphics[width=1.3cm]{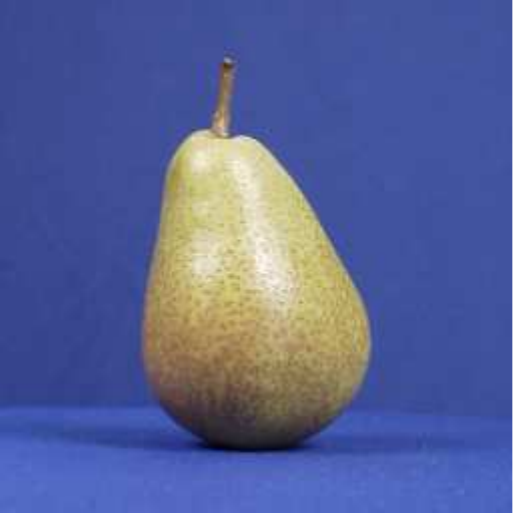}
  \caption{Images of two objects from the \acro{ETH-80} dataset. Each object has
    5 different views in our subset.}
  \label{fig:eth-sample}
\end{figure}

We first evaluated these techniques on a simple subset of the \acro{ETH-80} dataset \cite{eth80} to see basic properties of the performance.
This dataset contains 8 categories of objects, each of which has 10 individual objects; there are 41 images of each such object from different viewing angles.
We follow \textcite{Grauman07pyramid} in evaluating on a 400-image subset of the dataset, selecting 5 images per object that capture its appearance from different angles.
Sample images of two objects are shown in Fig.~\ref{fig:eth-sample}.

\begin{figure}[thb]
  \centering
  \includegraphics[width=8cm]{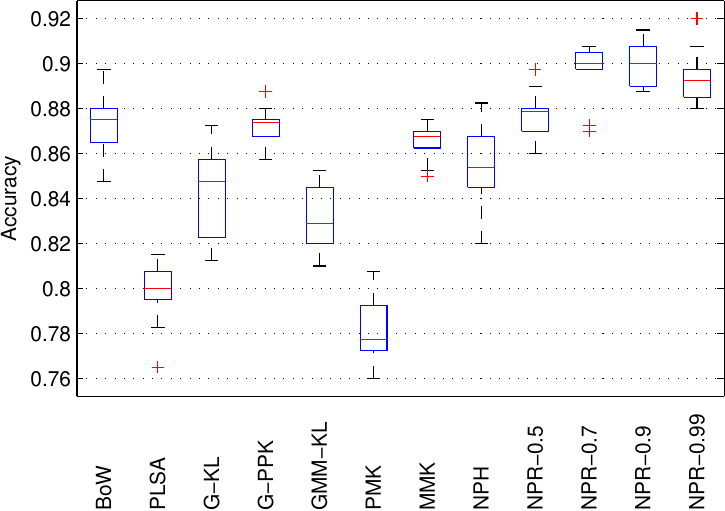}
  \caption{Classification accuracies on \acro{ETH-80}.}
  \label{fig:acc-eth}
\end{figure}

We used color \acro{SIFT} features for this dataset with bin size fixed at 6 pixels, as this problem does not require scale invariance.
We reduced the vectors' dimensionality to 18 with \acro{PCA}, preserving 50\% of the variance.
Each image is therefore represented by 576 18-dimensional pairs.

We tested the performance of classification into these 8 categories based on 2-fold cross-validation.
The results from $10$ random runs are shown in Fig.~\ref{fig:acc-eth}.

We can see that our R\'enyi-divergence kernels performed better than \acro{BoW}, and much better than the other methods.
In this test, \acro{BoW} achieved impressive results when properly tuned.
The improvement of \acro{NPR-0.9} (mean accuracy $89.93\%$, std dev $0.9\%$) over \acro{BoW} ($87.33\%$, std dev $1.4\%$) is statistically significant: a paired t-test shows a $p$-value of $2 \times 10^{-3}$. \acro{MMK} performs at the same level as \acro{BoW}, and also significantly worse than our methods.
It is also interesting to see that \acro{GMM}-based methods perform even worse than simple Gaussian-based methods.
This may be because it is harder to choose the parameters of a \acro{GMM}, or
because divergences between \acro{GMM}s could not be obtained precisely.
\acro{PMK} is not very accurate here, though it did evaluate quickly.

\begin{figure}[thb]
  \centering
  \includegraphics[width=8cm]{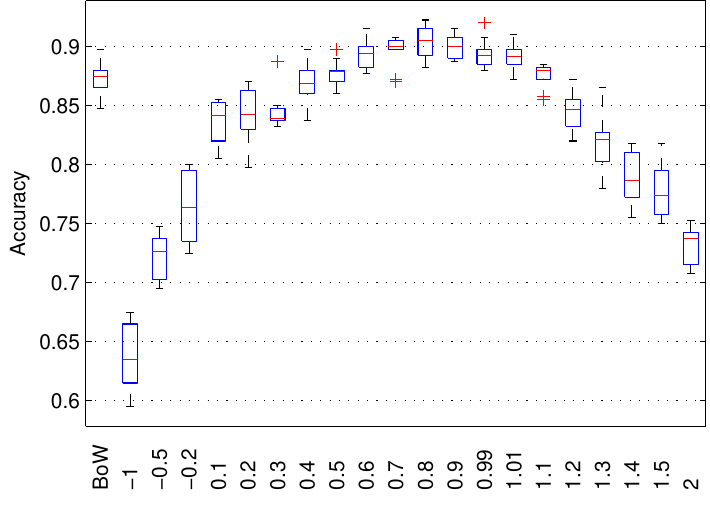}
  \caption{Classification accuracies on \acro{ETH-80} with R\'enyi-$\alpha$s for twenty $\alpha$s, as well as \acro{BoW}.}
  \label{fig:accs-eth-alphas}
\end{figure}

Fig.~\ref{fig:accs-eth-alphas} shows the performance of the R\'enyi-$\alpha$ kernel for many values of $\alpha$, along with the \acro{BoW} performance for context.
The best $\alpha$ values are clearly in the vicinity of 1, \ie near the \acro{KL} divergence,
though the performance seems to degrade more quickly when greater than 1 than when below.

\subsubsection{Scene Classification}

\begin{figure}[thb]
  \centering
  \includegraphics[height=1.9cm]{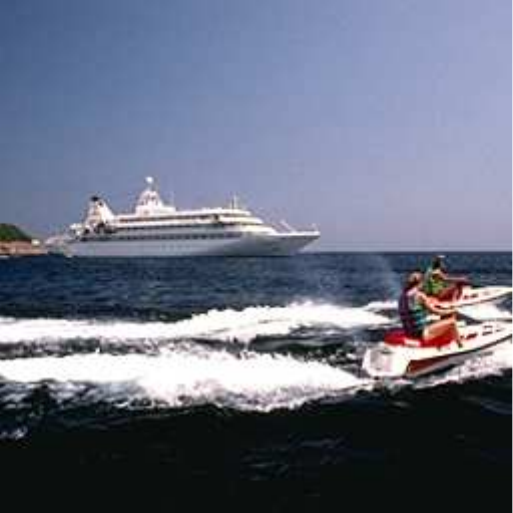}
  \includegraphics[height=1.9cm]{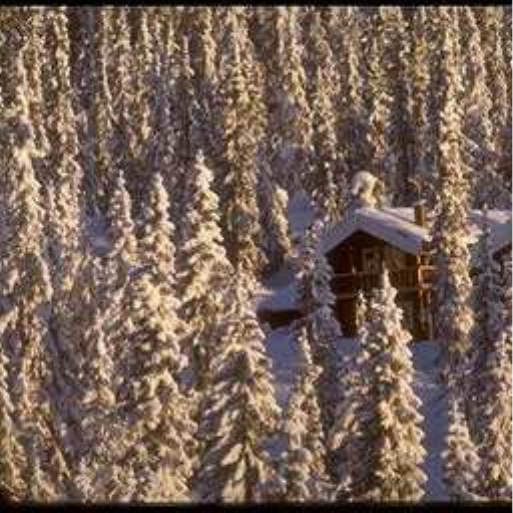}
  \includegraphics[height=1.9cm]{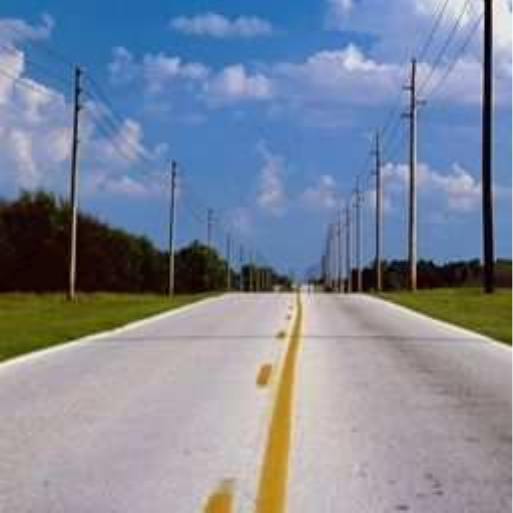}
  \includegraphics[height=1.9cm]{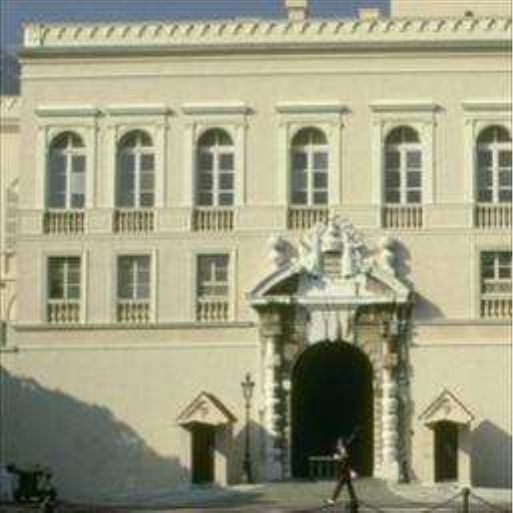}
  \includegraphics[height=1.9cm]{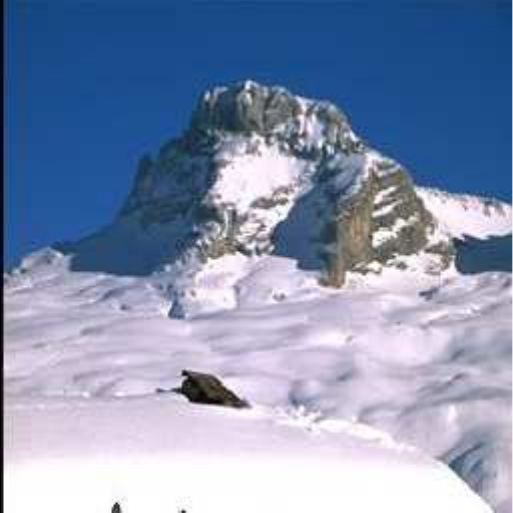}
  \includegraphics[height=1.9cm]{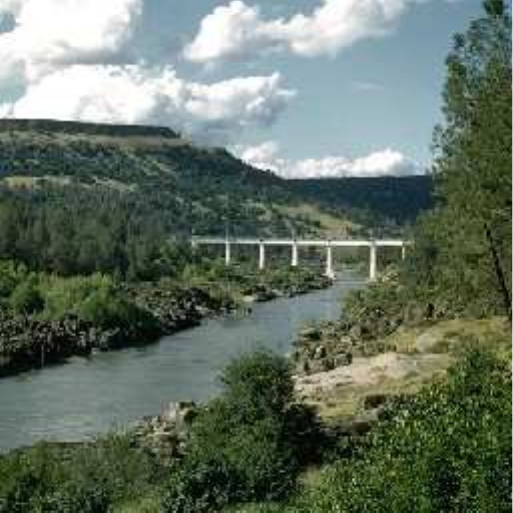}
  \includegraphics[height=1.9cm]{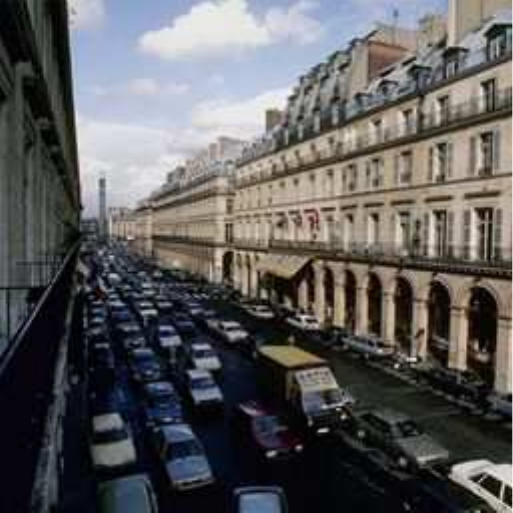}
  \includegraphics[height=1.9cm]{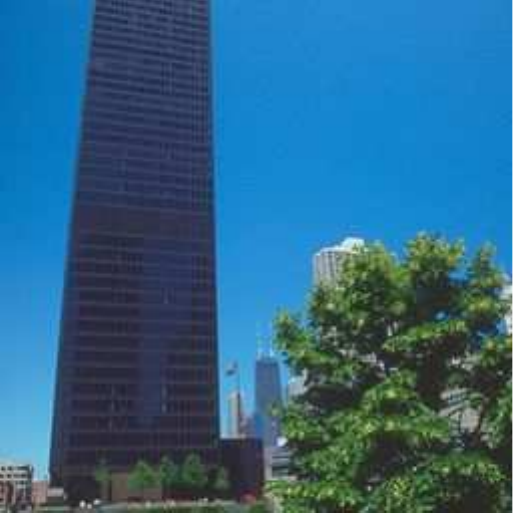}
  \caption{The 8 \acro{OT} categories:
    \emph{coast, forest, highway, inside city, mountain, open
      country, street, tall building}.}
  \label{fig:ot-sample}
\end{figure}

\begin{figure}[thb]
  \centering
  \includegraphics[width=8cm]{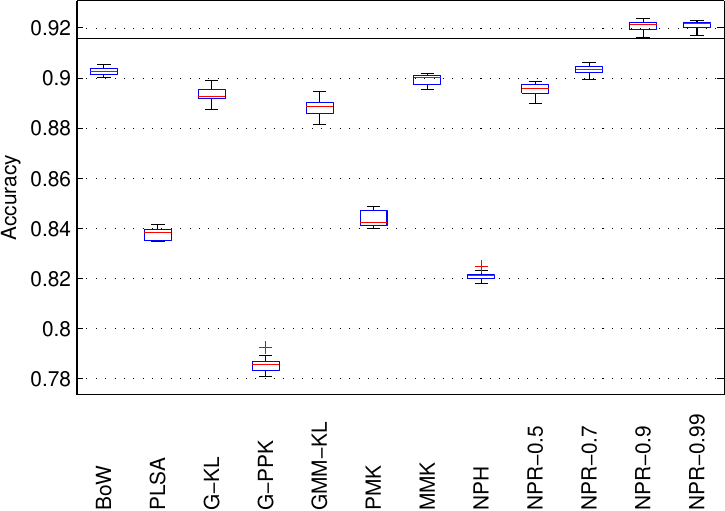}
  \caption{Accuracies on the \acro{OT} dataset; the horizontal line shows the best previously reported result \cite{context_color}.}
  \label{fig:acc-OT-cs}
\end{figure}

Scene classification using \acro{BoF}/\acro{BoW} representations is a well-studied problem for which many methods have been proposed \parencite{fei05BOW,bosch:08:bow,context_words}.
Here we test the performance of our nonparametric kernels against state-of-the-art methods.

We use the \acro{OT} dataset \parencite{dat_ot}, which contains $8$
outdoor scene categories:
\emph{coast, mountain, forest, open country, street, inside city, tall buildings, and highways}.
There are $2,688$ images in total, each about $256 \times 256$ pixels.
Sample images are shown in Fig.~\ref{fig:ot-sample}.
Our goal is to classify test images into one of the $8$ categories.

We used color \acro{SIFT} features, and also append the relative $y$ location of each patch (0 meaning the top of the image and 1 the bottom) onto the local feature vectors, allowing the use of some information about objects' locations in the images in classification.
(We chose not to include $x$ coordinates, because horizontal locations of objects generally carry little information in these scene images).
We used bin sizes of $\{6, 12, 18, 24, 30\}$ so that more global information can be captured.
A typical image therefore contains 1815 \acro{SIFT} feature vectors, each of dimensionality 384; these are reduced by \acro{PCA} to 53 dimensions (preserving 70\% of the variance) and then $y$ coordinates are appended.
Each dimension of the feature vectors was finally normalized to have zero mean and unit variance.

The accuracies of 10 random runs are shown in Fig.~\ref{fig:acc-OT-cs}.
Here results of 10-fold cross-validations are used so that we can directly compare to other published results.
\acro{GMM-PPK} is not shown because it is too low.

\acro{NPR}-0.99 achieved the best average accuracy of $92.11\%$, which is much better than \acro{BOW}'s $90.26\%$ (paired $t$-test $p = 1.4 \times 10^{-8}$).
Notably, this $92.11\%$ accuracy (std dev $0.18\%$) surpasses the best previous result of which we are aware, $91.57\%$~\parencite{context_color}.
For comparison, in 2-fold cross-validations the mean accuracies of \acro{NPR}-0.99 and \acro{BOW} are $90.85\%$ and $88.21\%$ respectively.

\subsubsection{Sport Event Classification}

These kernels can also be used for visual event classification \parencite{dat_sport} in the same manner as for scene classification.
We use the dataset from \textcite{dat_sport}, which contains Internet images of $8$ sport event categories:
\emph{badminton, bocce, croquet, polo, rock climbing, rowing, sailing,} and \emph{snowboarding}.
This dataset is considered more difficult than traditional scene classification, as it involves much more widely varying foreground activity than does \eg the \acro{OT} dataset.

\begin{figure}[htb]
  \centering
  \includegraphics[height=1.4cm]{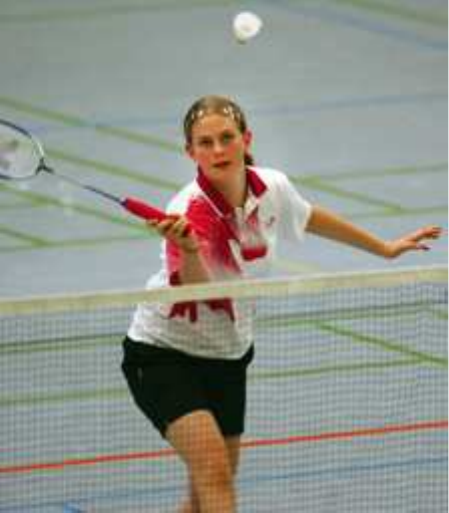}
  \includegraphics[height=1.4cm]{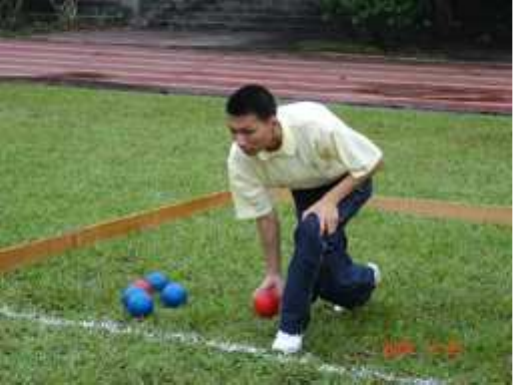}
  \includegraphics[height=1.4cm]{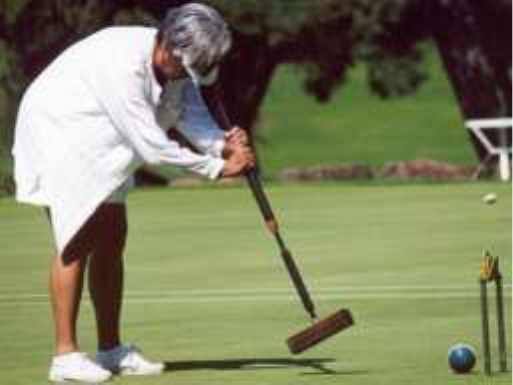}
  \includegraphics[height=1.4cm]{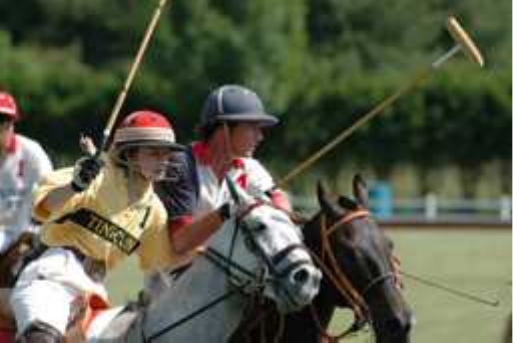}
  \includegraphics[height=1.4cm]{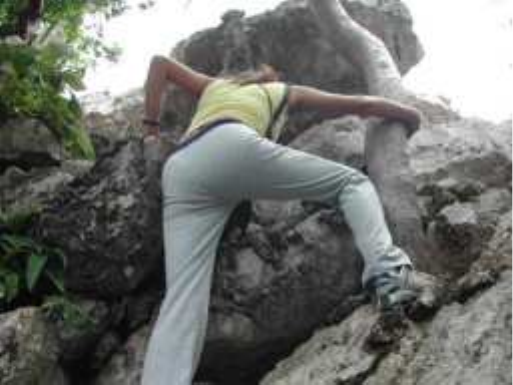}
  \includegraphics[height=1.4cm]{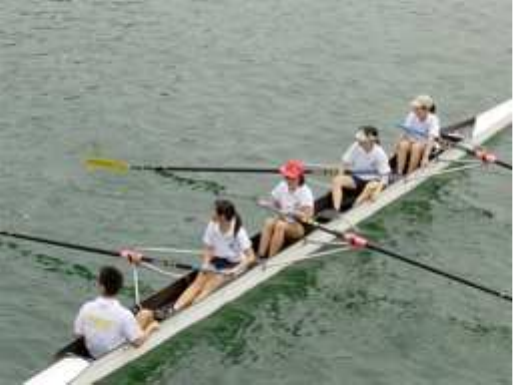}
  \includegraphics[height=1.4cm]{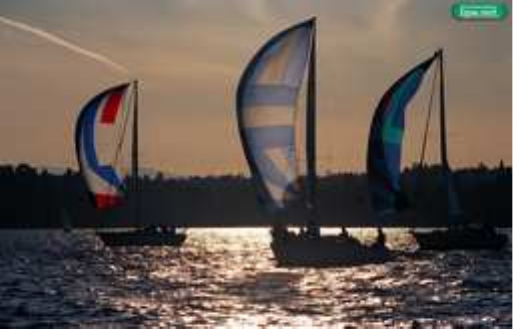}
  \includegraphics[height=1.4cm]{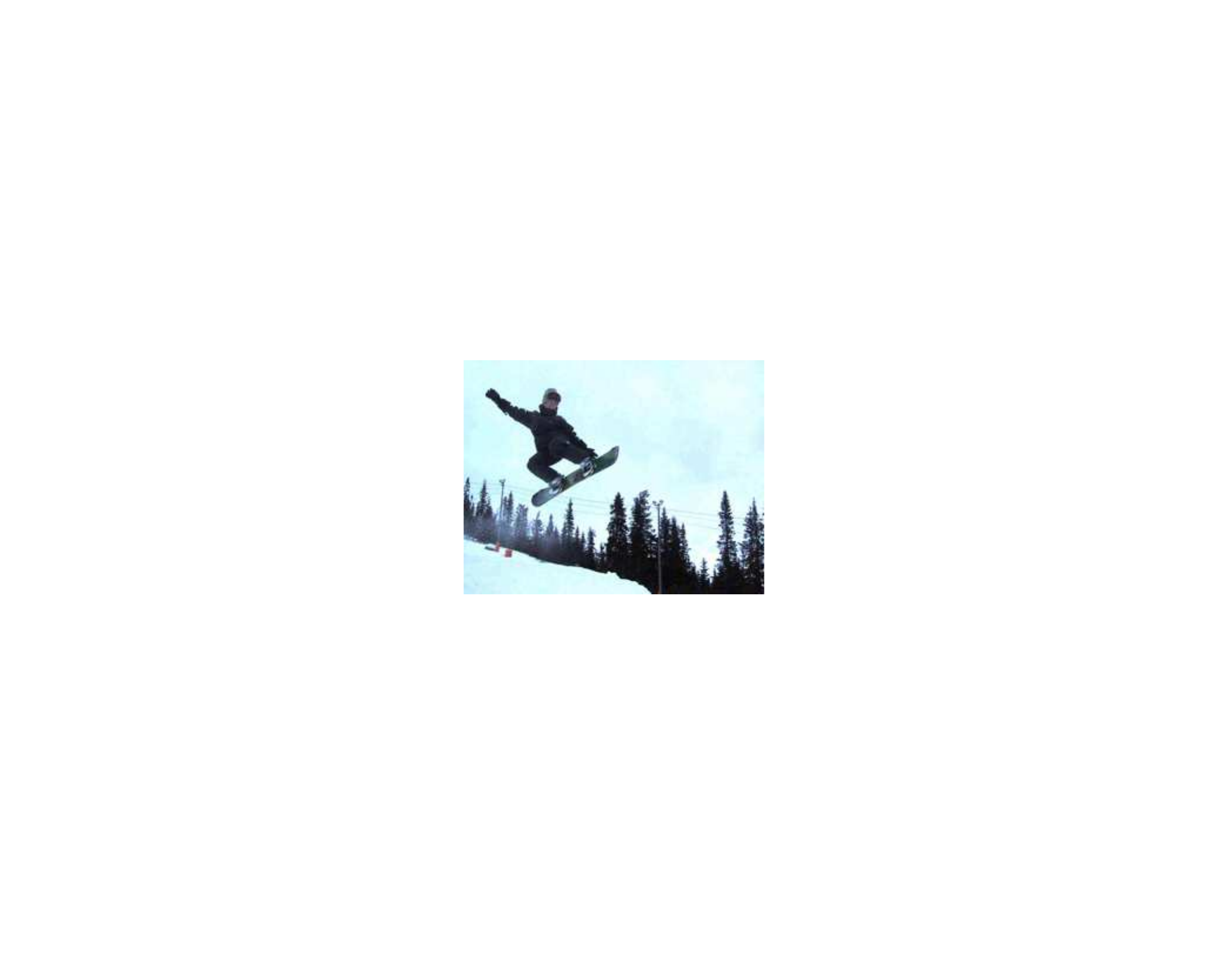}
  \caption{The 8 sports: badminton, bocce, croquet,
    polo, rock climbing, rowing, sailing, snowboarding.}
  \label{fig:sport-sample}
\end{figure}

Following \textcite{dat_sport}, we used the first $130$ images from each category.
For each image, we extracted color \acro{SIFT} features with radii of $\{6, 9, 12\}$ and reduced their dimensionality to $57$ (preserving 70\% of the variance).
As image sizes vary, each \acro{BoF} group contains between $295$ and $1542$ feature vectors.
To incorporate spatial information, we included the patches' relative $x$ and $y$ locations in the feature vectors, and normalize each dimension as before.

\begin{figure}[thb]
  \centering
  \includegraphics[width=8cm]{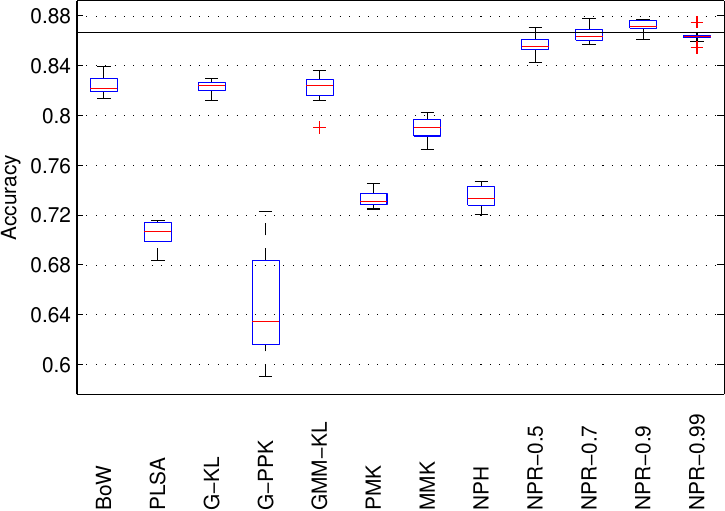}
  \caption{Classification accuracy on the sport dataset, with the best previously reported result \parencite{sparsecode}.}
  \label{fig:acc-sport}
\end{figure}

Fig.~\ref{fig:acc-sport} shows the accuracies of 10 random 2-fold
cross-validations.
Our R\'enyi-$.9$ kernel again achieved the best accuracy, with $87.18\%$ (std dev $0.46\%$).
This performance is one standard deviation above the reported result for state-of-the-art methods such as those of \textcite{sparsecode}, which attained $86.7\%$.
It is worth noting that these methods achieved significant performance increases by learning features, whereas we used only \acro{PCA} \acro{SIFT}.
Compared to previous results, we can see that the performance of \acro{PPK} and \acro{MMK} methods decreased (we did not show \acro{GMM-PPK} here because its accuracy is too low).
The \acro{BoW} method, though worse than R\'enyi-$.9$ with 82.42\% ($p = 6 \times 10^{-9}$), again performs well, showing its wide applicability.

\subsection{Turbulence Data}\label{section:turbulence}

\begin{figure*}[tb]
    \centering
    \subfigure[Classification probabilities]{
        \includegraphics[width=3in]{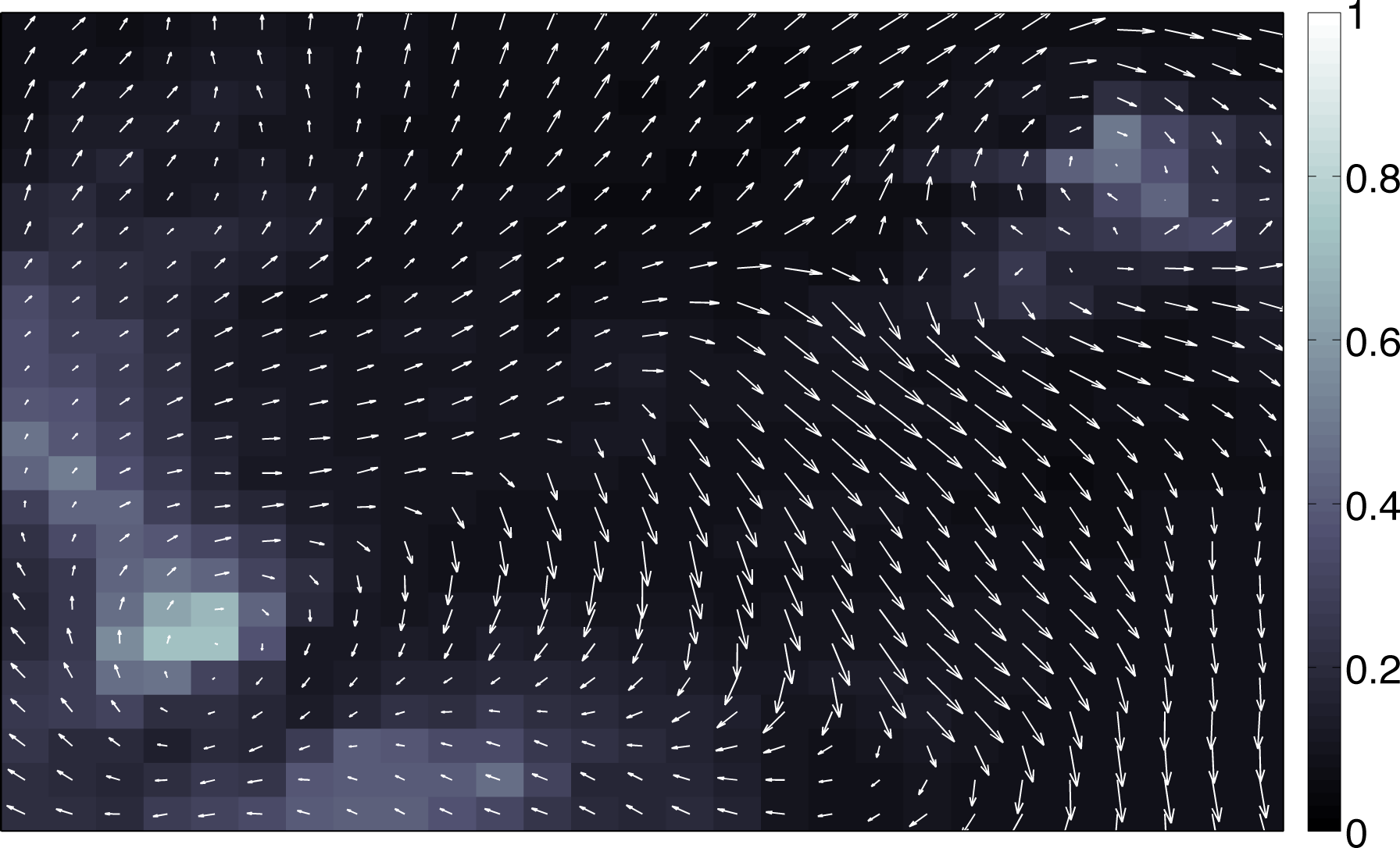}
        \label{fig:turb-scores}
    }
    \subfigure[Anomaly scores]{
        \includegraphics[width=3in]{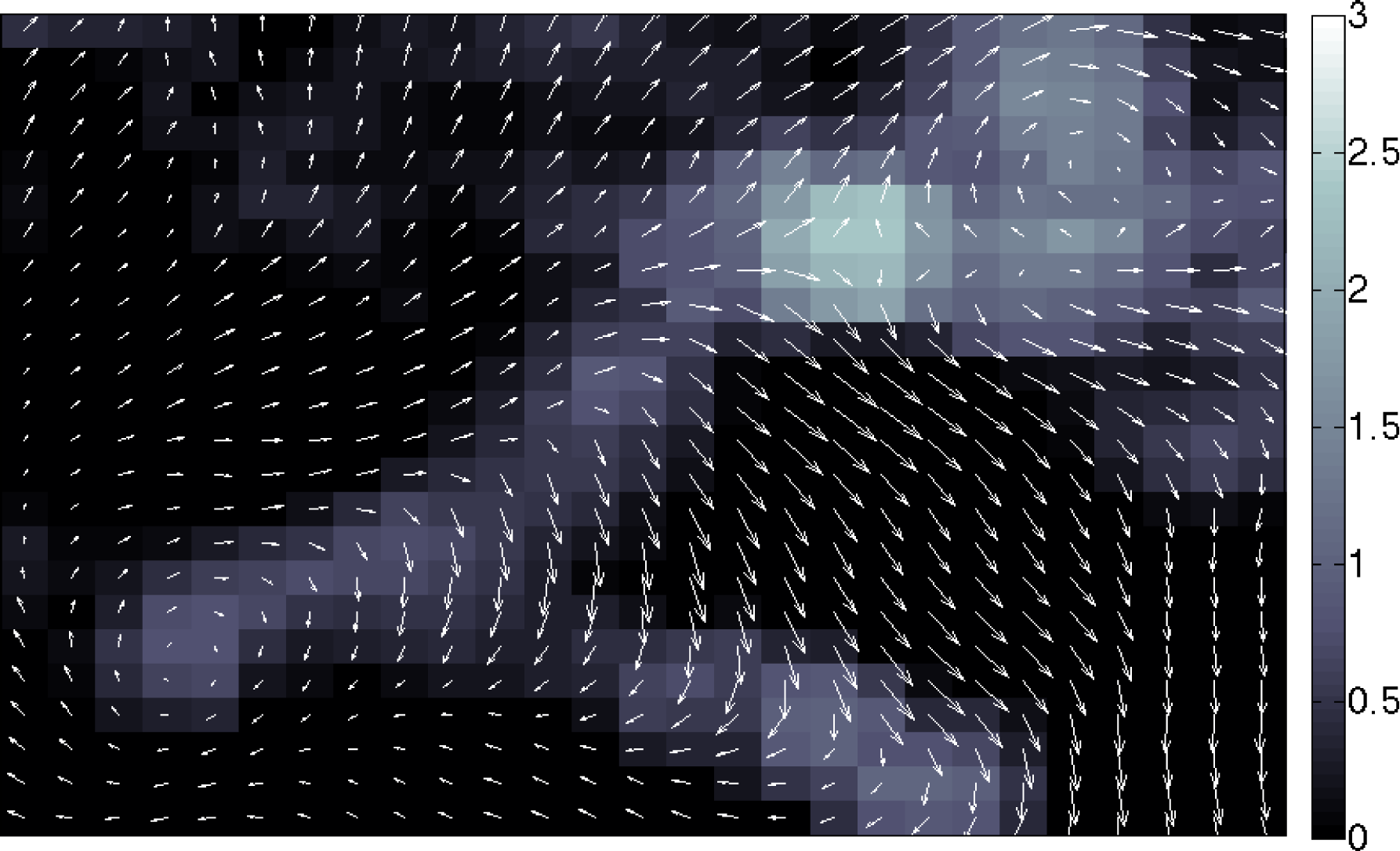}
        \label{fig:turb-anomaly}
    }
    \caption{Classification and anomaly scores along with velocities for one $54\times48$ slice of the turbulence data.}
    \label{fig:turb-score-quivers}
\end{figure*}

One of the challenges of doing science with modern large-scale simulations is identifying interesting phenomena in the results, finding them, and computing basic statistics about when and where they occurred.

We present the results of exploratory experiments on using the proposed kernel to assist in this process, using turbulence data from the JHU Turbulence Data Cluster\footnote{\httpurl{turbulence.pha.jhu.edu}} (TDC, \parencite{perlman:07:turb}).
TDC simulates fluid flow through time on a 3-dimensional grid, calculating 3-dimensional velocities and pressures of the fluid at each step.
We used one time step of a contiguous $256\times256\times128$ sub-grid.

\subsubsection{Classification}

We first consider the task of finding one particular class of interesting phenomena in the data: stationary vortices parallel to the $xy$ plane.
Distributions are defined on the $x$ and $y$ components of each velocity in an $11\times11$ square, along with that point's squared distance from the center.
The latter feature was included due to the intuition that velocity in a vortex is related to its distance from the center of the vortex.

\begin{figure}[h]
    \centering
    \subfigure[Positive]{
        \includegraphics[width=1in]{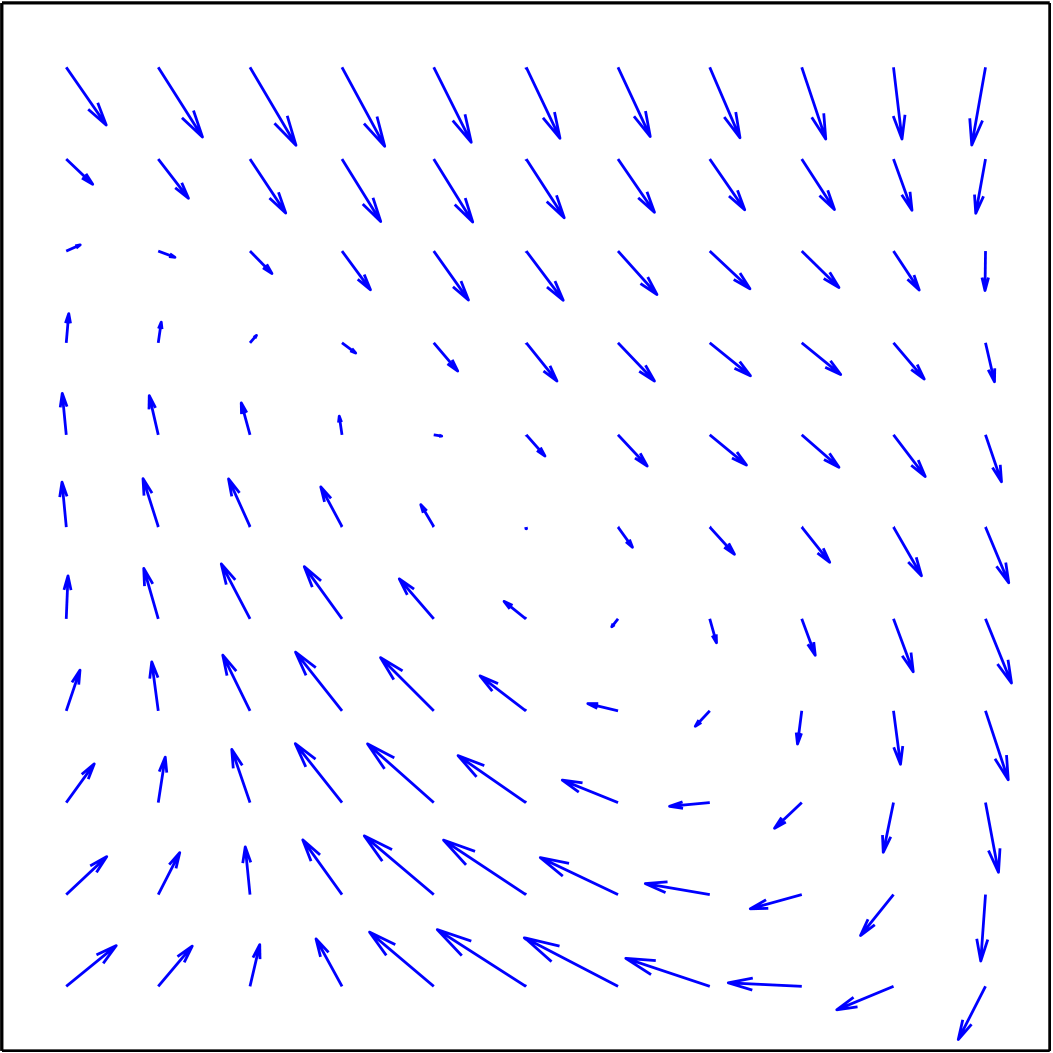}
        \label{fig:turb-train-vortex} 
    }
    \subfigure[Negative]{
        \includegraphics[width=1in]{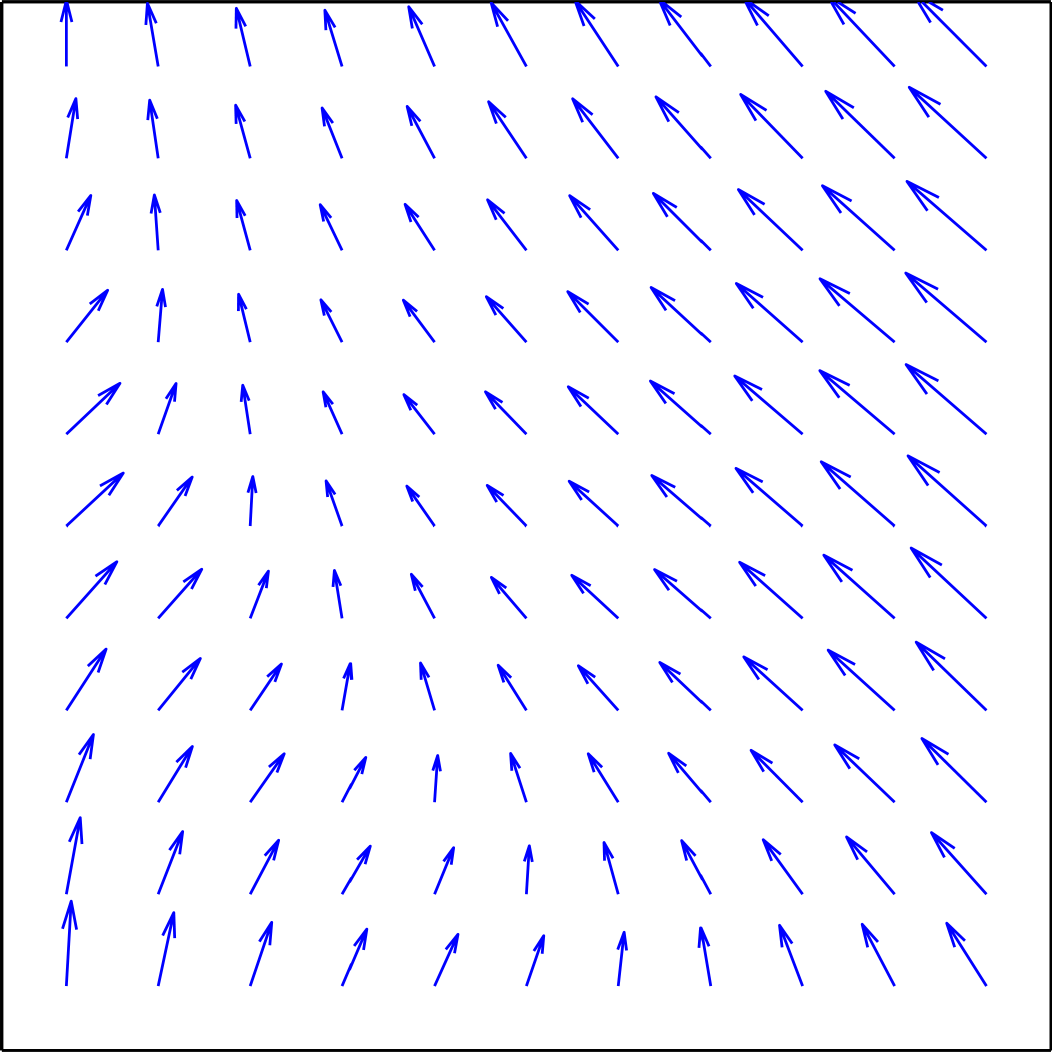}
        \label{fig:turb-train-simpleneg} 
    }
    \subfigure[Negative]{
        \includegraphics[width=1in]{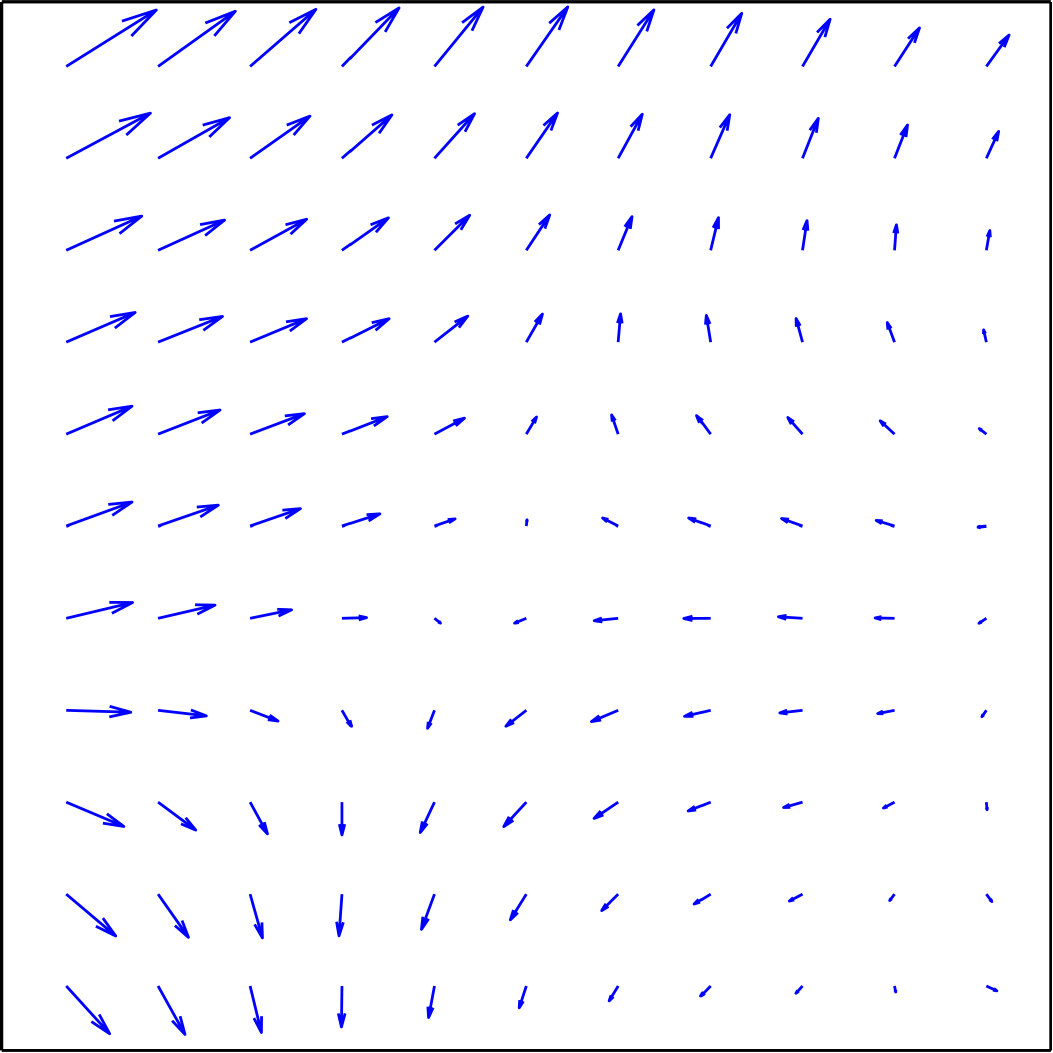}
        \label{fig:turb-train-diamondneg} 
    }
    \caption{Training examples for the vortex classifier.}
    \label{fig:turb-examples}
\end{figure}

We trained an SVM on a manually-labeled training set of 11 positives and 20 negatives, using a Gaussian kernel based on Hellinger distance.
Some representative training examples are shown in Fig.~\ref{fig:turb-examples}.
On this small training set, leave-one-out cross-validation scores varied significantly based on the random outcome of parameter tuning.
In 1000 independent runs, Renyi-$.9$ kernels achieved mean accuracy $79\%$, Hellinger $82\%$, and $L_2$ $84\%$; each had a standard deviation of $5\%$.

A classifier based on Hellinger kernels was then used inductively to evaluate groups along $z$-slices of the data, with a grid resolution of $2\times2$.
One slice of the resulting probability estimates is shown in Fig.~\ref{fig:turb-scores}; the arrows represent the mean velocity at each classification point.
The high-probability region on the left is a canonical vortex, while the slightly-lower probability region in the upper-right deviates a little from the canonical form.
Note that some other areas show somewhat complex velocity patterns but mostly have low probabilities.


\subsubsection{Anomaly Detection}
As well as finding instances of known patterns, part of the process of exploring the results of a large-scale simulation is seeking out unexpected phenomena.
To demonstrate anomaly detection with our methods, we trained a one-class SVM on 100 distributions from the turbulence data, with centers chosen
randomly.
Its evaluation on the same region as Fig.~\ref{fig:turb-scores} is
shown in Fig.~\ref{fig:turb-anomaly}.
The two vortices are picked out, but the area with the highest score is a diamond-like velocity pattern, similar to Fig.~\ref{fig:turb-train-diamondneg}; these may well be less common in the dataset than are vortices.

We believe that the proposed classifiers and anomaly detectors serve as a proof of
concept for a simulation exploration tool that would allow scientists to
iteratively look for anomalous phenomena and label some of them.
Classifiers could then find more instances of those phenomena and compute statistics about their occurrence, while anomaly detection would be iteratively refined to highlight only what is truly new.

\subsection{Regression with Scalar Response}
We now turn to the problem of distribution regression with scalar response (Section~\ref{sec:drsr}), showing examples of how it can be used for learning real-valued functionals of distributions from samples in a nonparametric way.

\begin{figure}[b] 
    \centering
    \hbox{\subfigure[Skewness of Beta]{
        \includegraphics[height = 1.1in]{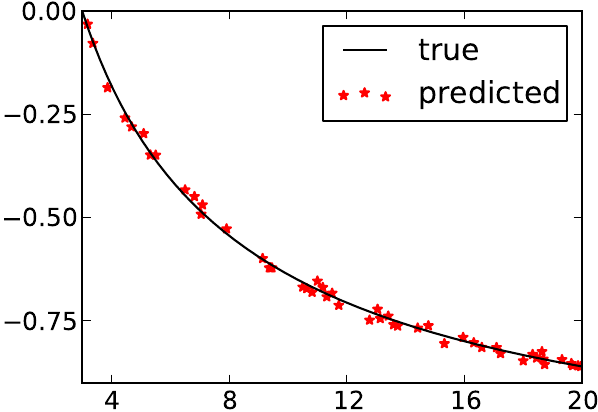}
        \label{fig:beta-skewness}}
    \subfigure[Entropy of Gaussian]{
        \includegraphics[height = 1.1in]{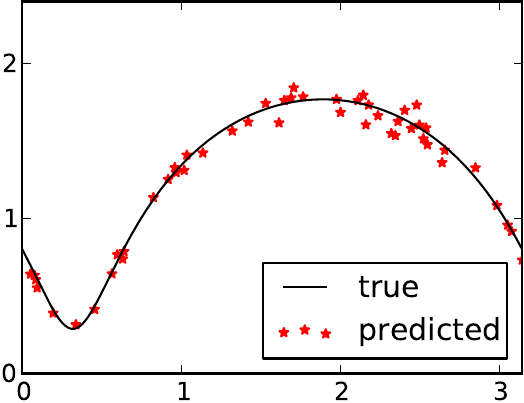}
        \label{fig:rotated-2Dgaussian}}}
    \caption{(a) Learned skewness of $Beta(a,3)$ distributions as a function of $a \in [3,20]$. (b) Learned entropy of a 1d marginal distribution of a rotated 2d Gaussian distribution as a function of the rotation angle in $[0,\pi]$.}
\end{figure}

\begin{figure*}[t!]
    \centering
    \subfigure[Rotated 2d Gaussians]{
        \includegraphics[width = 2in]{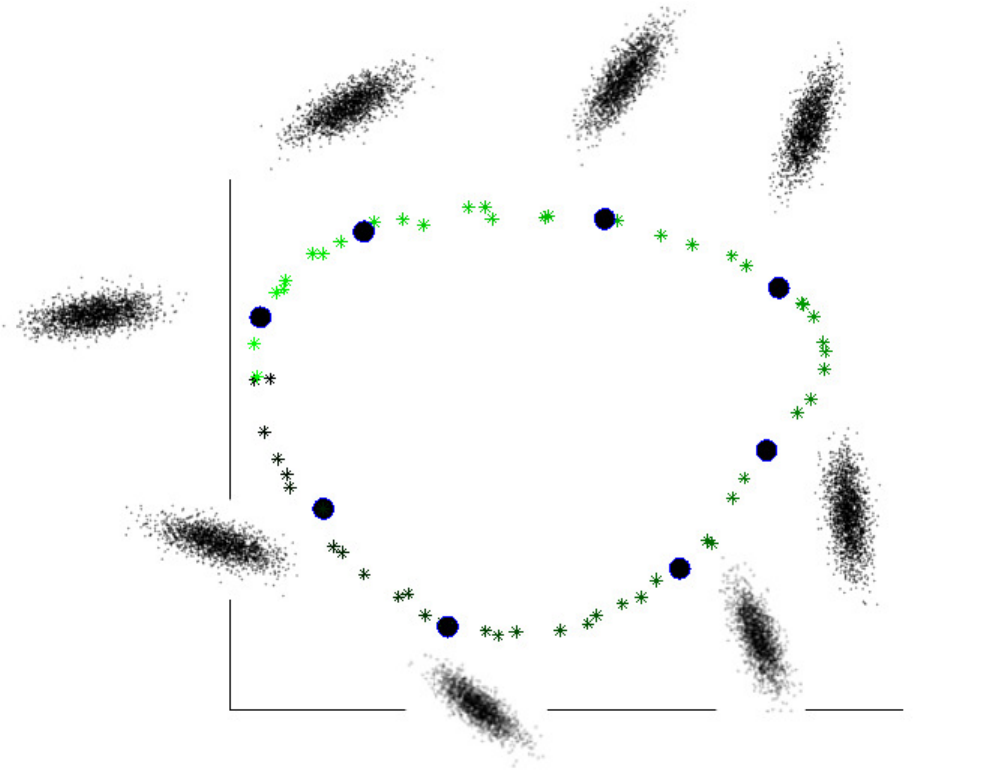}
        \label{fig:rotatedgaussian}}
    \subfigure[Euclidean image distances]{
        \includegraphics[width= 2in]{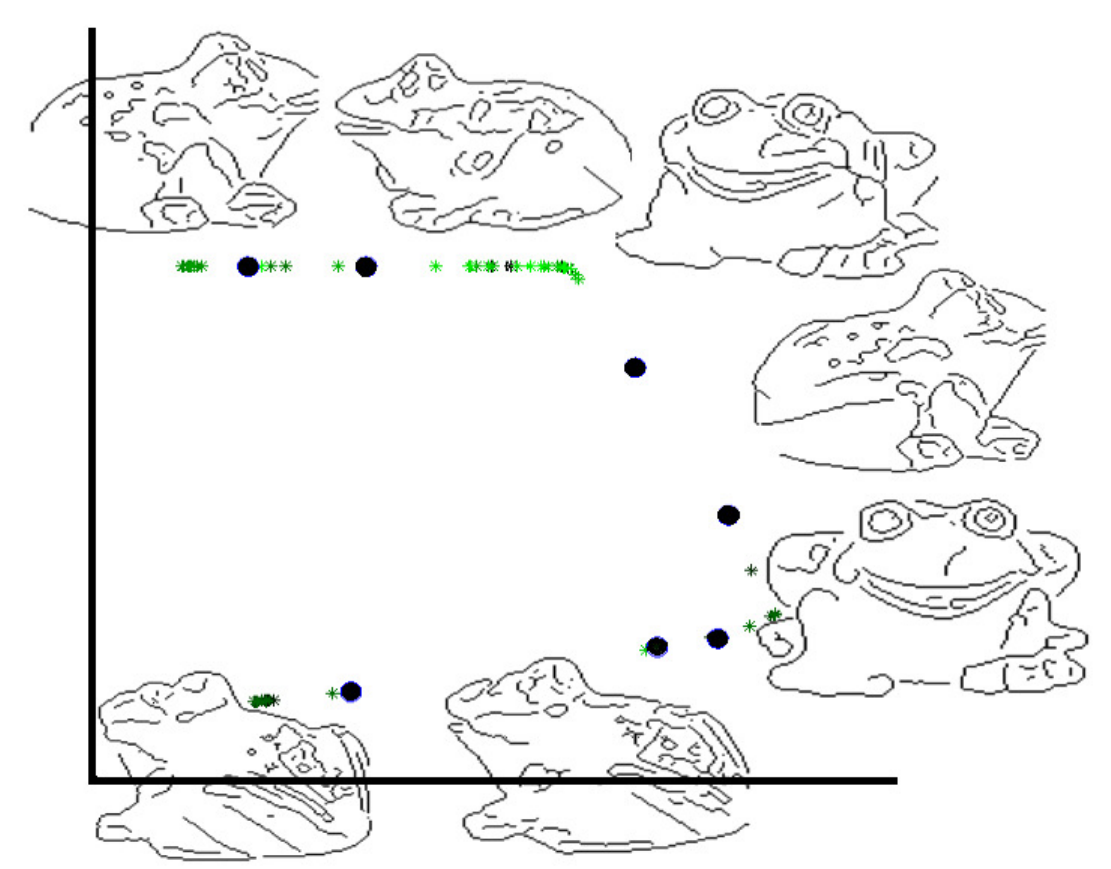}
        \label{fig:coilresults-imgdist}
    }
    \subfigure[Kernel estimation]{
        \includegraphics[width= 2in]{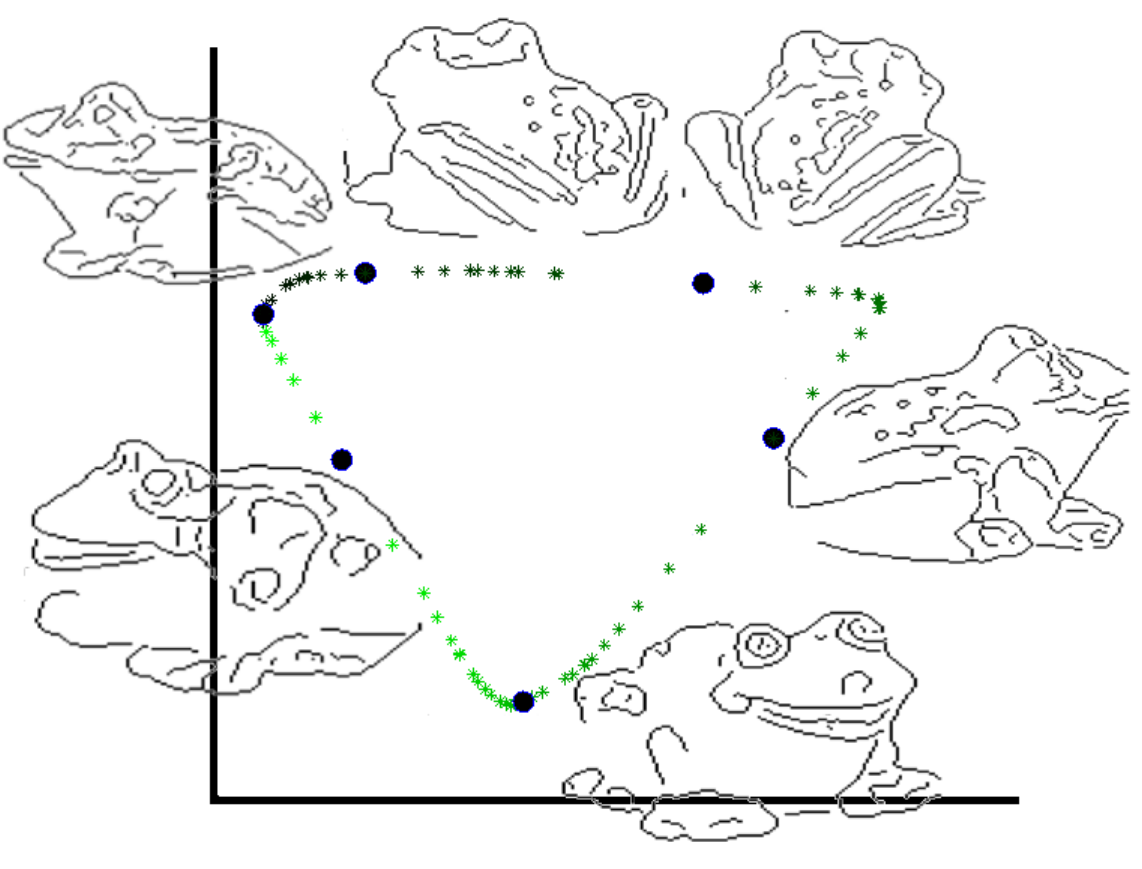}
        \label{fig:coilresults-distrib}
    }
    \caption{
        \subref{fig:rotatedgaussian} LLE of rotated 2d Gaussian distributions.
        \subref{fig:coilresults-imgdist} LLE using Euclidean distances between edge-detected images. The embedding failed to keep the local geometry of the original pictures.
        \subref{fig:coilresults-distrib} LLE considering the edges as samples from unknown distributions. Nearby objects are successfully embedded to nearby places.
    }
\end{figure*}

We first generated 350 \iid sample sets of size 500 from $Beta(a,3)$ distributions, where $a$ was randomly sampled from $[3,20]$, and used 300 sample sets for training and 50 for testing.
Our goal was to learn the skewness of $Beta(a, b)$ distributions,
whose true value is
\begin{equation*}
    s = \frac{2 (b-a) \sqrt{a+b+1}}{(a+b+2) \sqrt{ab}}
\end{equation*}
but that we will approximate using the labeled sample sets.
Fig.~\ref{fig:beta-skewness} displays the predicted values for the 50 test sample sets.
In this experiment we used a R\'enyi-$.9$ kernel,
with regression parameter $\varepsilon = 0.01$
and $C$ and $\sigma$ tuned as in classification.
The testing RMSE was $0.012$.

In the next experiment, we learned the entropy of Gaussian distributions.
We generated a $2 \times 2$ covariance matrix $\Sigma = C C^T$
by drawing each element of $C \in \R^{2 \times 2}$ from $\mathcal{N}(0, 1)$,
obtaining $\Sigma = [0.29$ $-0.57$; $-0.57$ $1.83]$.
We then generated 300 sample sets of size 500
with covariance matrices defined by rotating $\Sigma$:
\begin{gather*}
    \{\mathcal{N}(0, M) \}_{i=1}^{150},
    \text{ with } M = R(\alpha_i) \, \Sigma \, R^T(\alpha_i) \in\mathbb{R}^{2 \times 2}
\end{gather*}
where $R(\alpha_i)$ is a 2d rotation matrix
with rotation angle $\alpha_i = i \pi / 150$.
Our goal was to learn the entropy of the first marginal distribution
    $H = \frac{1}{2} \ln(2 \pi \, e \, M_{1,1})$.
Fig.~\ref{fig:rotated-2Dgaussian} shows the learned entropies of the 50 test sample sets.
We used the same settings as for the beta skewness;
the testing RMSE was $0.058$.

\subsection{Locally Linear Embedding}

Here we show results on locally linear embedding of distributions.
This algorithm uses the linear DRDR method as a subroutine.

We generated 2000 \iid sample points from each of 63 rotated versions of a Gaussian distribution,
with mean zero and covariance matrix $R(\alpha_i) \, \Sigma \, R(\alpha_i)^T$.
Here $\Sigma = [9 \; 0; \; 0 \; 1]$,
$i = 1, \dots, 63$,
and $R(\alpha_i)$ denotes the 2d rotation matrix with rotation angle $\alpha_i=(i-1)/20$.
We ran the LLE algorithm (Section~\ref{sec:drdr-lle}) and embedded these distributions into 2d.
The results are shown in Fig.~\ref{fig:rotatedgaussian}.
We can see that our method preserves the local geometry of these sample sets: distributions with similar rotation angles are mapped into nearby points.

\begin{figure}  
    \centering
    \subfigure[Original image]{
        \includegraphics[width = 1.2in]{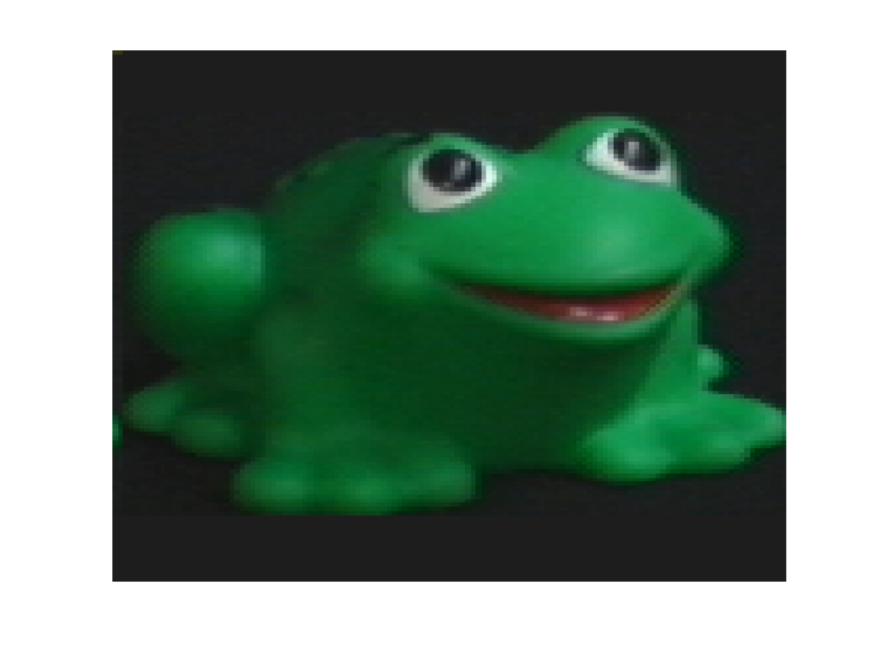}
        \label{fig:coildataset-orig}}
    \subfigure[Edge-detected]{
        \includegraphics[width = 1.2in]{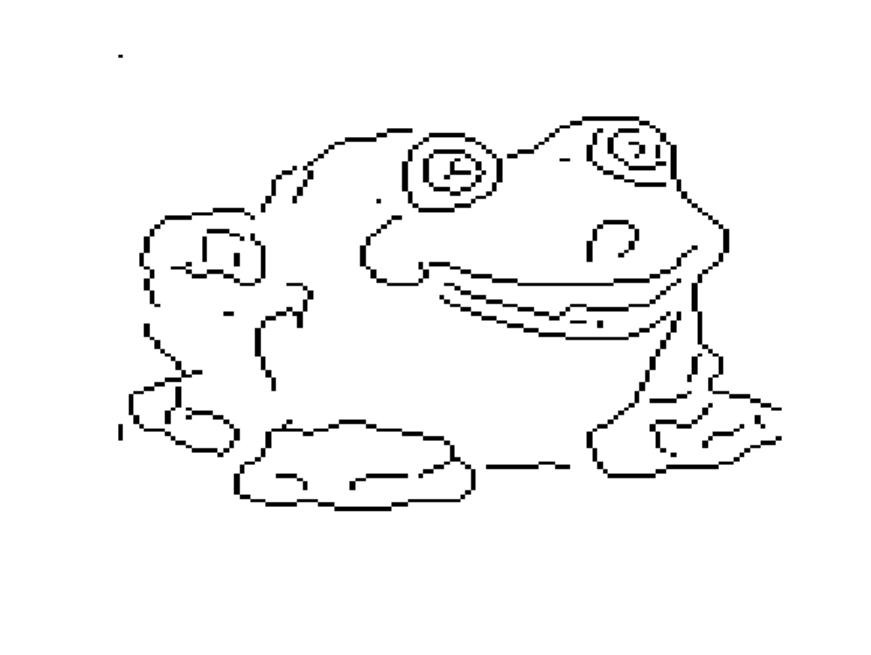}
        \label{fig:coildataset-edge}}
    \caption{An example view of the COIL object.}
    \label{fig:coildataset}
\end{figure}

We repeated this experiment on the edge-detected images of an object in the COIL dataset.%
\footnote{\httpurl{www.cs.columbia.edu/CAVE/software/softlib/coil-100}}
We converted the 72 128x128 color pictures of a rotated 3D object
to grayscale and performed Canny edge detection on them,
as shown in Fig.~\ref{fig:coildataset}.
The number of detected edge points on these images was between 845 and 1158.

Our goal is to embed these edge-detected images into a 2d space preserving proximity. This problem is easy using the original images, but challenging when only the edge-detected images are available.
If we simply use the Euclidean distances between these edge-detected images, the standard LLE algorithm fails, as shown in Fig.~\ref{fig:coilresults-imgdist}. 
Considering the edge-detected images as sample points from unknown 2d distributions, however, the LLE algorithm on distributions yields a successful embedding, shown in Fig.~\ref{fig:coilresults-distrib}.
The embedded points preserve proximity and local geometry of the original images.

\section{Discussion and Conclusion}
We have posed the problem of performing machine learning on groups of data points as one of machine learning on distributions, where the data points are viewed as samples from an unknown underlying distribution for the group.
We proposed a kernel on distributions and provided nonparametric methods for consistently estimating those kernels based on sample sets.
We further demonstrated that our methods work well across a range of supervised and unsupervised tasks, matching or surpassing state-of-the-art performance on several well-studied image classification tasks and showing promising results in other areas.

\printbibliography

\clearpage
\appendix

We give here an overview of the estimator defined in Section~\ref{section:Kernel-estimation}, which is a generalization of the work by \textcite{poczos11uai}.

\subsection*{$k$-NN Based Density Estimators} \label{section:knn-density-estim}

$k$-\acro{NN} density estimators operate using distances between
the observations in a given sample and their $k$th nearest neighbors.
Although we \emph{do not perform consistent density estimation}, the ideas of these
estimators are crucial in deriving the divergence estimator.

Let $X_{1:n}\doteq(X_1,\ldots,X_n)$ be an
\iid sample from a distribution with density $p$, and similarly let
$Y_{1:m}\doteq(Y_1,\ldots,Y_m)$ be an \iid sample from a
distribution having density $q$.
Let $\rho_k(i)$ denote the Euclidean distance of the $k$th nearest
neighbor of $X_i$ in the sample $X_{1:n} \setminus \{X_i\}$, and similarly let
$\nu_k(i)$ denote the distance of the $k$th nearest neighbor of
$X_i$ in the sample $Y_{1:m}$. Let $\bar{c}_d$
denote the volume of a $d$-dimensional unit ball.
\textcite{Loftsgaarden65nonparametric} define the $k$-\acro{NN}
based density estimators of $p$ and $q$ at $X_i$ as
\begin{equation*}
 \hat{p}_k(X_i) = \frac{k}{(n-1) \, \bar{c}_d \, \rho_k^d(i)}, \quad
 \hat{q}_k(X_i) = \frac{k}{m \, \bar{c}_d \, \nu_k^d(i)}.
\end{equation*}
Note that these estimators are consistent only when
$k(n) \to \infty$, as claimed in \ref{thm:knn-1}.
We will use these density estimators in our proposed divergence
estimators, but \emph{we will keep $k$ fixed} and still be able to
prove their consistency.
\begin{theorem}[convergence in probability]\label{thm:knn-1}
If $k(n)$ denotes the number of neighbors applied at sample size $n$,
$\lim_{n\to\infty} k(n)=\infty$, and $\lim_{n\to\infty}
n/k(n)= \infty$, then $\hat{p}_{k(n)}(x) \to_{p} p(x)$ for almost all
$x$.
\end{theorem}

\subsection*{Consistency of $\widehat D_{\alpha,\beta}(X_{1:n}\|Y_{1:m})$}
If $\M$ is the support of $p$, our goal is to estimate \eqref{eq:Dalpha-beta}:
\begin{align*}
    D_{\alpha,\beta}(p\|q) = \int_{\M} p^\alpha(x) \, q^{\beta}(x) \, p(x) \, \ud x.
\end{align*}
Our proposed estimator \eqref{eq:D_hat} is equivalent to
\begin{align*}
    \widehat{D}_{\alpha,\beta}
      = \frac{B_{k,d,\alpha,\beta}}{n}
      \sum_{i=1}^n \left( (n-1) \, \rho_k^{d}(i) \right) ^{-\alpha}
      \; \left( m \, \nu_k^{d}(i) \right) ^{-\beta}, \label{eq:D_hat}
\end{align*}
where
$B_{k,d,\alpha,\beta}\doteq \bar{c}_d^{-\alpha-\beta} \frac{\Gamma(k)^2}{\Gamma(k-\alpha)\Gamma(k-\beta)}$
and $\bar{c}_d$ is the volume of a $d$-dimensional unit ball.

Let $\B{(x,R)}$ denote a closed ball
around $x \in \mathbb{R}^d$ with radius $R$, and let
$\Vol\bigl(\B({x,R})\bigr)=\bar{c}_dR^d$ be its volume.
In the following theorems we will assume that almost all points of $\M$ are in
its interior and that $\M$ has the following additional property:
\begin{equation*}
    r_{\M} \doteq
    \inf_{0<\delta<1} \inf_{x \in \M}
    \frac{\Vol\bigl(\B({x,\delta}\bigr)\cap\M)}%
         {\Vol\bigl(\B({x,\delta})\bigr)}
    > 0.
\end{equation*}
If $\M$ is a finite union of bounded convex sets, then this condition holds.

\begin{theorem}[Asymptotic unbiasedness] \label{thm:asympt-unbiased}
Let $-k<\alpha,\beta<k$. If $0<\alpha<k$, then let $p$ be bounded away
from zero and uniformly continuous; when $-k < \alpha < 0$, let $p$ be bounded. Similarly,
if $0<\beta<k$, then let $q$ be bounded away
from zero and uniformly continuous; if $-k < \beta < 0$, then let $q$ be bounded. Under these conditions we have that
\begin{equation*}
\lim_{n,m\to \infty} \Exp\left[ \widehat D_{\alpha,\beta}(X_{1:n}\|Y_{1:m})\right] = D_{\alpha,\beta}(p\|q),
\end{equation*}
\ie the estimator is asymptotically unbiased.
\end{theorem}

The following theorem provides conditions under which $\widehat{D}_{\alpha,\beta}$ is
$L_2$ consistent.  In the previous theorem we have stated conditions
that lead to asymptotically unbiased divergence estimation. In
the following theorem we will assume that the estimator is
asymptotically unbiased for $(\alpha,\beta)$ as well as
for $(2\alpha,2\beta)$, and also assume that
$D_{\alpha,\beta}(p\|q)<\infty,D_{2\alpha,2\beta}(p\|q)<\infty$.

\begin{theorem}[$L_2$ consistency] \label{thm:L2-consistent}
Let $k \geq 2$ and $-(k-1)/2<\alpha,\beta<(k-1)/2$. If $0<\alpha<(k-1)/2$, then let $p$ be bounded away
from zero and uniformly continuous; if $-(k-1)/2<\alpha<0$, then let $p$ be bounded. Similarly,
if $0<\beta<(k-1)/2$, then let $q$ be bounded away
from zero and uniformly continuous; if $-(k-1)/2<\beta<0$, then let $q$ be bounded. Under these conditions we have
  \begin{equation*}
\lim_{n,m\to \infty} \Exp\left[ \left(\widehat D_{\alpha,\beta}(X_{1:n}\|Y_{1:m})-D_{\alpha,\beta}(p\|q)\right)^2\right]=0;
\end{equation*}
that is, the estimator is $L_2$ consistent.
\end{theorem}

\subsubsection*{Proof Outline for Theorems~\ref{thm:asympt-unbiased}-\ref{thm:L2-consistent}} \label{section:proofs-outline}

We can repeat the argument of \textcite{poczos11aistats}.
Using the $k$-\acro{NN} density estimator, we can estimate $1/p(x)$ by
$n \bar{c}_d \rho_k^d(x) / k$.
From the Lebesgue lemma, one can prove that the distribution of $n \bar{c}\rho_k^d(x)$ converges weakly to an Erlang distribution with mean $k/p(x)$, and variance $k/p^2(x)$ \parencite{Leonenko-Pronzato-Savani2008}.
In turn, if we divide $n \bar{c}_d \rho_k^d(x)$ by $k$, then asymptotically it has mean $1/p(x)$ and variance $1/(kp^2(x))$.
This implies that indeed (in accordance with Theorem~\ref{thm:knn-1}) $k$ should diverge in order to get a consistent estimator; otherwise, the variance will not disappear.
On the other hand, $k$ cannot grow too fast: if, say, $k=n$, then the estimator would be simply $\bar{c}\rho_k^d(x)$, which is a useless estimator since it is asymptotically zero whenever $x\in\supp(p)$.

Luckily, in our case we do not need to apply consistent density estimators. The trick is that \eqref{eq:Dalpha-beta} has a special form:
  $\int p(x) p^\alpha(x) q^\beta(x) \ud x$.
Our estimator \eqref{eq:D_hat} is (rearranging some terms):
  \begin{align} \label{eq:oneterm-only}
      \frac{\Gamma(k)^2 \, k^{-\alpha-\beta}}{\Gamma(k-\alpha)\,\Gamma(k-\beta)}\;
      \frac{1}{n}
      \sum_{i=1}^n \left(\hat p_k(X_i)\right)^{\alpha}
                \left(\hat q_k(X_i)\right)^{\beta},
\end{align}
where the leftmost factor is a correction that ensures asymptotic unbiasedness. Using the Lebesgue lemma again, we can prove that the distributions of $\hat p_k(X_i)$ and $\hat q_k(X_i)$ converge weakly to the Erlang distribution with means $k/p(X_i)$, $k/q(X_i)$ and variances $k/p^2(X_i)$, $k/q^2(X_i)$, respectively \parencite{Leonenko-Pronzato-Savani2008}. Furthermore,
they are conditionally independent for a given $X_i$. Therefore, ``in the limit'' \eqref{eq:oneterm-only} is simply the
empirical average of the products of the $\alpha$th and $\beta$th powers of independent Erlang distributed variables. These moments can be calculated in closed form. For a fixed $k$, the $k$-\acro{NN} density estimator is not consistent since its variance does not vanish. In our case, however, this variance will disappear thanks to the empirical average in \eqref{eq:oneterm-only} and the law of large numbers.

While the underlying ideas of this proof are simple, there are a couple of serious gaps in it. Most importantly, from the Lebesgue lemma we can guarantee only the weak convergence of $\hat p_k(X_i)$, $\hat q_k(X_i)$ to the Erlang distribution. From this weak convergence we cannot imply that the moments of the random variables converge too. To handle this issue, we will need stronger tools such as the concept of asymptotically uniformly integrable random variables \parencite{vanderVaart07asymptotic}, and we also need the uniform generalization of the Lebesgue lemma. As a result, we need to put some extra conditions on the densities $p$ and $q$ in Theorems~\ref{thm:asympt-unbiased}--\ref{thm:L2-consistent}.
The details follow from a slight generalization of the derivations in \textcite{poczos11aistats}.

%
%
%
%
%
%
%
\end{document}